\documentclass[10pt,journal,compsoc]{IEEEtran}
\usepackage{amsmath}
\usepackage{mathrsfs}
\usepackage{bbm}
\usepackage{graphicx}
\usepackage{subfigure}
\usepackage{mathrsfs}
\usepackage{multirow}
\usepackage{balance} 
\usepackage[vlined,boxed,commentsnumbered, ruled, linesnumbered]{algorithm2e}
\usepackage{amsfonts,amssymb}
\usepackage{array}
\usepackage{setspace} 
\usepackage{xcolor}
\usepackage{soul}

\ifCLASSOPTIONcompsoc
 \usepackage[nocompress]{cite}
\else
 \usepackage{cite}
\fi
\ifCLASSINFOpdf
\else
\fi
\begin{document}
%
\title{\huge Adversarial Detection by Latent Style Transformations}
%
%
%
%
\author{Shuo~Wang,~\IEEEmembership{Member,~IEEE,}
		Surya~Nepal,~\IEEEmembership{Member,~IEEE,}
		Alsharif~Abuadbba,~\IEEEmembership{Member,~IEEE,}
		Carsten~Rudolph,~\IEEEmembership{Member,~IEEE,}
		 and~ Marthie~Grobler,~\IEEEmembership{Member,~IEEE}
\IEEEcompsocitemizethanks{\IEEEcompsocthanksitem Shuo Wang, Surya Nepal, Alsharif Abuadbba and Marthie Grobler are with CSIRO's Data61 \& Cybersecurity CRC, Australia.\protect\\
E-mail: shuo.wang@monash.edu
\IEEEcompsocthanksitem Carsten Rudolph is with Faculty of Information Technology at Monash University, Melbourne, Australia.\protect\\
E-mail: Carsten.Rudolph@monash.edu.
}
}
%
%

\markboth{IEEE Transactions on , August 2020}%
{Wang \MakeLowercase{\textit{et al.}}: Adversarial Detection by Latent Style Transformations}
%



\IEEEtitleabstractindextext{%
\begin{abstract}
\textcolor{black}{
Detection-based defense approaches are effective against adversarial attacks without compromising the structure of the protected model. However, they could be bypassed by stronger adversarial attacks and are limited in their ability to handle high-fidelity images.
In this paper, we explore an effective detection-based defense against adversarial attacks on images (including high-resolution images) by extending the investigation beyond a single-instance perspective to incorporate its transformations as well. 
Our intuition is that the essential characteristics of a valid image are generally not affected by non-essential style transformations, for example, a slight variation in the facial expression of a portrait would not alter its identification.
In contrast, adversarial examples are designed to affect only a single instance at a time, with unpredictable effects on a set of transformations of the instance. 
Consequently, we leverage a controllable generative mechanism to conduct the non-essential style transformations for a given image via modification along the style axis in the latent space. 
Next, the consistency of prediction between the given input and its style transformations is used to distinguish adversarial instances. 
Based on experiments on three image datasets, including high-resolution images, we demonstrated that our defense could detect 90-100 percent of adversarial examples produced by various state-of-the-art adversarial attacks, with a low false-positive rate. 
}
\end{abstract}
}

\maketitle
\IEEEdisplaynontitleabstractindextext
%
\IEEEpeerreviewmaketitle
\IEEEraisesectionheading{\section{Introduction}\label{sec:introduction}}
\textcolor{black}{
DNNs have shown vulnerabilities to adversarial attacks, specifically, misclassification of adversarial examples. 
It is possible for an attacker to generate adversarial examples by adding tiny and indistinguishable perturbations to legitimate (clean) samples \cite{goodfellow6572explaining}. 
Detecting approaches have been demonstrated to be effective to improve the robustness of DNN models against adversarial attacks, such as Defense-GAN \cite{samangouei2018defense}, MagNet \cite{meng2017magnet}, FBGAN \cite{bao2018featurized} and Image Transformation-based detection, which requires no modification to the protected classifier. 
Existing detection defenses, however, suffer from three significant challenges. 
First, they directly investigate the pixel space of a single instance, such as by comparing the pixel-wise reconstruction errors between the original instance and its reconstruction copy. Therefore, adversarial attacks capable of introducing tiny pixel perturbations may bypass the detection approaches. 
Second, they are limited in their ability to handle images with high resolution due to the complexity involved in reconstruction or perturbation. 
Third, existing transformation detection approaches are commonly conducted on deterministic transformations in pixel space, such as fixed angle rotations, which are typically attack-specific and are only valid for a particular dataset. 
}

\textcolor{black}{
To address these challenges, this paper proposes an effective detection-based defense scheme against strong adversarial attacks by extending the investigation beyond a single instance perspective to incorporate its transformations on specific styles as well. 
Our intuition is that the essential characteristics of a valid image are generally consistent with non-essential style transformations, e.g., a slight change in the facial expression of a human portrait would not affect identification. 
In contrast, adversarial examples are designed to affect only one instance (single point) at a time through unstructured adversarial perturbations (random pixel noise). 
The impact of the adversarial perturbation may not be predictable or stable on the transformations (multi-point) of that instance. 
}

\textcolor{black}{
Consequently, we leverage a controllable generative mechanism to conduct the non-essential style transformations, called in\underline{V}ertible \underline{A}utoencoder based on the \underline{S}tyleGAN2 generator via \underline{A}dversarial learning (VASA). 
Specifically, the given image could be encoded into the representation in the latent space (known as latent codes) of the generator of VASA, where data augmentation could be controlled along a specified style axis in the latent space by modifying the latent codes. 
The prediction consistency between the input and its augmented copies is then used to distinguish adversarial instances. 
Our contributions are summarized as follows: 
(1) To the best of our knowledge, our defense is the first attempt to overcome the limitations of investigation from a single instance perspective and to develop an effective detection-based defense against adversarial attacks using controllable style transformations. 
(2) We propose VASA for efficiently encoding images into the latent space of the generator of a powerful generative model, and then conducting controllable style transformations based on latent shifting along a specified style axis in the latent space. Additionally, we incorporate randomness and automation into the transformation process.
(3) We provide an efficient adversarial detection that detects suspicious inputs based on the consistency of classification results between an image and its edited copies with non-essential style transformations, without introducing any extra training or knowledge of the process to generate adversarial examples. Besides, we also explore detection strategies under white-box adversaries. 
(4) We perform extensive evaluations using different detection strategies against attacks with black-box and white-box knowledge about the detector on three real-world datasets, including 1024$\times$1024 high-resolution face images. Our results suggest that our defense can achieve $90\%$ to $100\%$ detecting success rates against different state-of-the-art attacks, with a low false-positive rate. Besides, the false-positive error could be further reduced by the reconstruction. 
}

\section{Detection Defense: Key Insights}
\textcolor{black}{
The adversarial example generally aims to design specific pixel noise for each input to conduct the misleading prediction. On the other hand, detection-based defenses commonly investigate each input to find some suspicious patterns within the pixel space of a single instance. 
As these two parties are both operating in pixel space from the perspective of a single instance, one side always beats the other, as a Min-Max zero-sum game. 
Therefore, powerful attacks could always be designed to undermine the current defense strategy. 
Accordingly, the research question (RQ) is \textit{How can adversarial examples be more efficiently detected by extending the investigating scope from a single point to a broader one? } For example, the detection analyzes not only the single instance but also a set of its transformations.
}

\begin{figure}[!htb]
	\centering
	\setlength{\abovecaptionskip}{-0.05cm}
	\setlength{\belowcaptionskip}{-0.2cm}
	\subfigure[Illustration of style transformation on face image. Top: horizontal angle. Middle: eye opening state. Bottom: smiling expression.]{
		\includegraphics[width=3.0in]{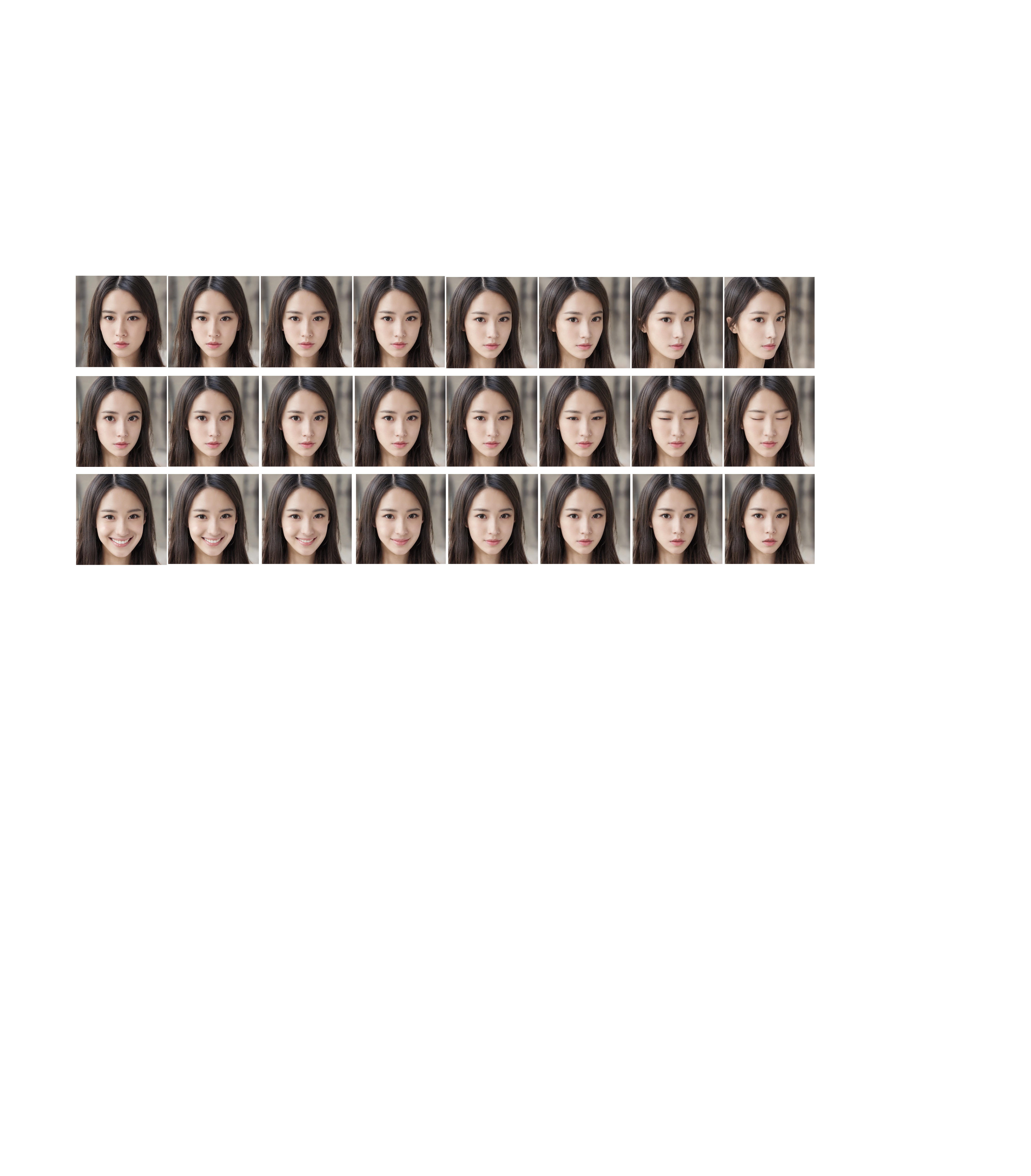}  
		
	}
	\subfigure[Illustration of style transformation on LSUN (cat and car).]{    \includegraphics[width=3.0in]{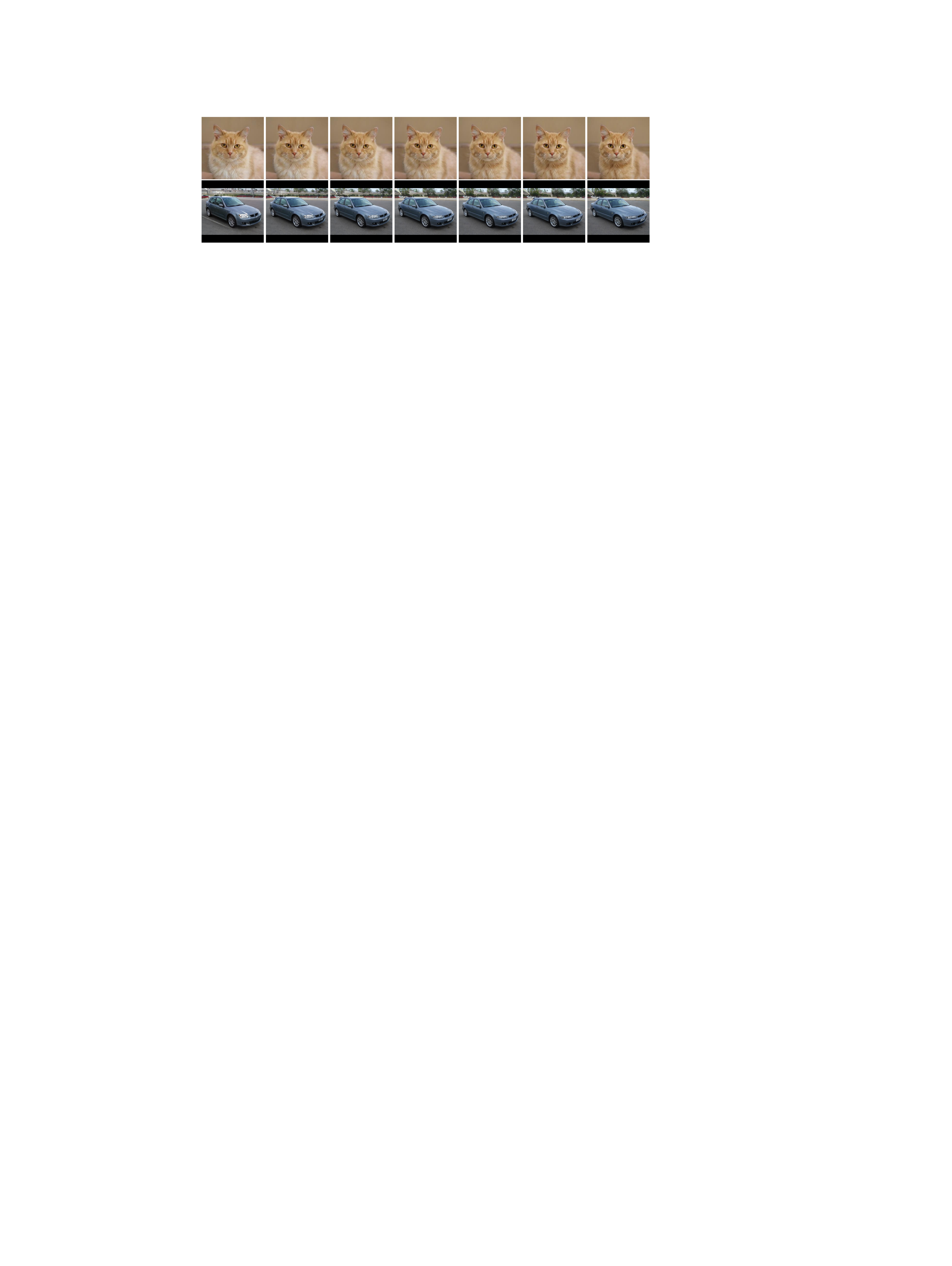}     }
	\caption{Illustration of the style transformations. }
\label{fig:demo}
\end{figure}

\textcolor{black}{\textbf{Intuition.} In general, 
the essential content of a legitimate image is resistant to image transformations in non-essential styles.
Human portraits, for example, share similar facial expression semantic patterns (non-essential styles), but differ in terms of identification information (essential characteristics). 
Fig. \ref{fig:demo} (a) demonstrates the non-essential style transformations for the same human portrait, including the angle of the face, the state of the eyes, and the smile.
Fig. \ref{fig:demo} (b) illustrates the style transformations of LSUN images of a cat (fur color) and a car (headlight style). 
Therefore, the identification predictions between a given human portrait image and its transformations with slight changes in facial expression tend to be consistent.  
Conversely, adversarial attacks are designed and performed in the pixel space for a single instance at a time, resulting in an unclear effect on its various style transformations. 
The reason is that the unstructured adversarial perturbation for a targeted instance does not follow the distribution of legitimate instances, resulting in unpredictable and unstable distortions in prediction after conducting style transformations. 
Consequently, our work suggests that the detection approach could examine the combination of each processed data point and a set of its style transformation copies, leading to a powerful scope-enhanced detection system.
In the ideal scenario, the detection would also be attack-independent. 
Accordingly, we employ generative models to apply style transformations to the input, seeking to identify distinguishable patterns of style transformations between legitimate and attacked samples in an effective manner. 
}

\begin{figure*}[!htb]
	\centering
	\setlength{\abovecaptionskip}{-0.05cm}
	\setlength{\belowcaptionskip}{-0.2cm}
	\subfigure[Illustration of latent shifting and reconstruction. The red line is the decision boundary of two classes (White and Black dots). The black triangle is the adversarial example for the white dot.]{    \includegraphics[width=5.0in]{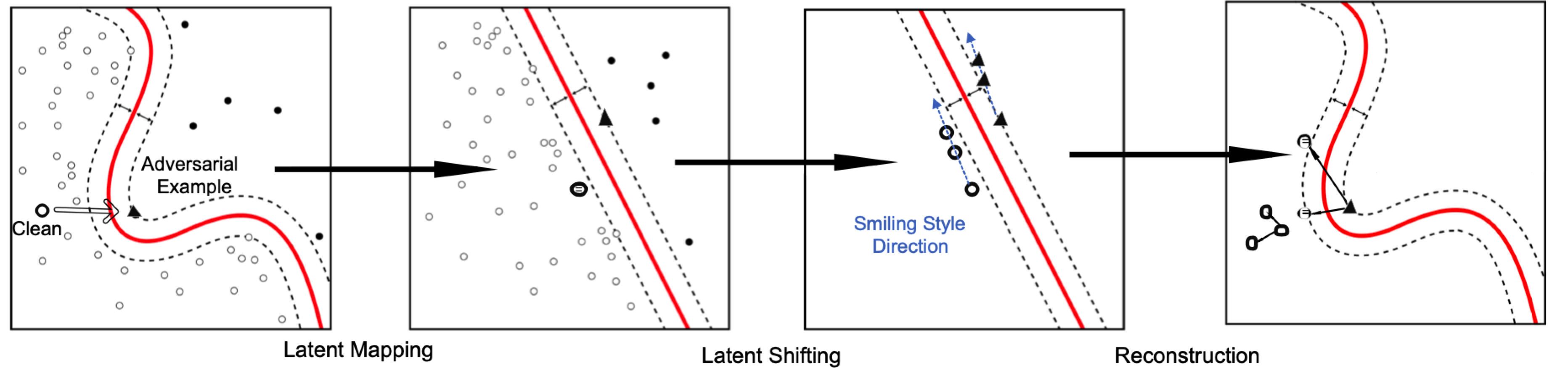}  }
	\subfigure[Demonstration of our defense using hand-written digits.]{    \includegraphics[width=5.0in]{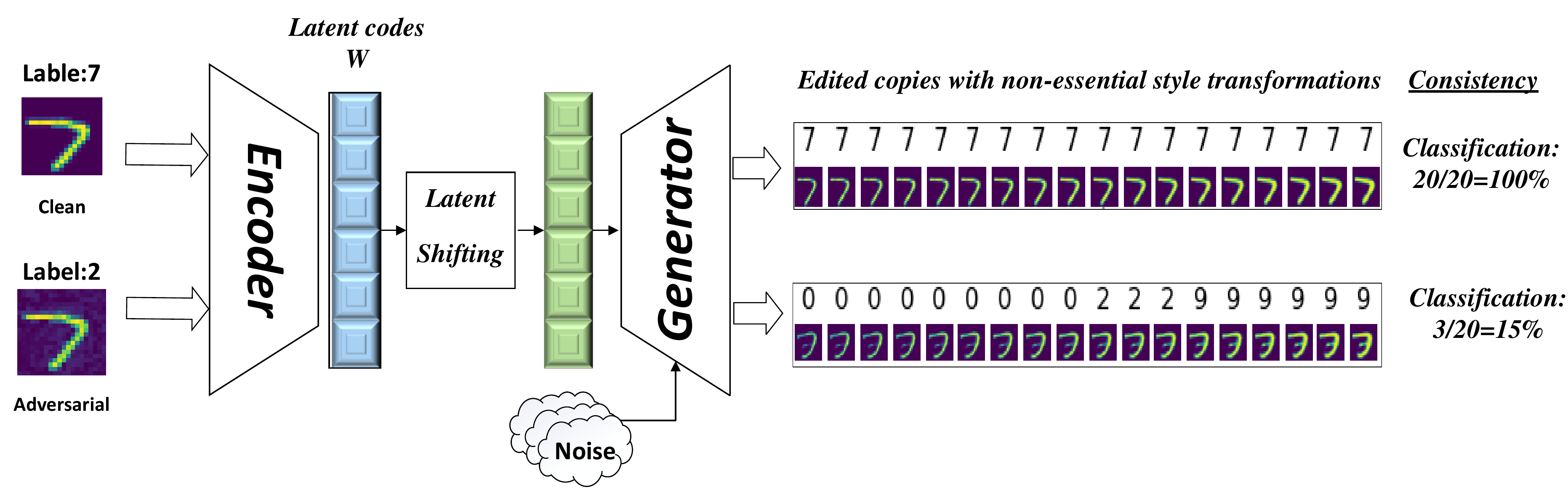}     }
	\caption{Illustration of the non-essential style transformations and VASA-defense. }
	\label{fig:illustrate}
\end{figure*}

\textcolor{black}{
We then demonstrate how latent shifting and reconstruction via generative models can be used to design such a detection mechanism systematically. 
Specifically, we use a controllable generative scheme VASA to conduct the non-essential style transformations, consisting of an encoder and a generator.
Then, we utilize the encoder to map the image to a representation in the latent space (latent codes) of the StyleGAN2 generator through adversarial learning, as shown in the second subfigure of Fig. \ref{fig:illustrate}(a). The reversed latent codes are specialized to be hierarchical and disentangled.
The next step is to identify the correspondence between non-essential styles and latent codes and identify the style axis within the latent space, such as the smiling style axis recognized in Fig. \ref{fig:illustrate}(a). 
The latent codes are then shifted along the latent style axis using linear interpolation and are then reconstructed using the generator, producing a set of edited copies with specific style transformations, e.g., controlling the thickness of hand-written digits. 
It is demonstrated in Fig. \ref{fig:illustrate} (b) that the classification results of legitimate instances are consistent with their edited copies with non-essential style transformations. 
In contrast, the unstructured adversarial perturbation of adversarial examples cannot follow the manifold of legitimate instances, resulting in an unexpected distortion in the prediction of the edited copies. 
In terms of classification-based consistency, we conclude that the small latent shift along with the non-essential style axis slightly affects the essential characteristics (e.g., identification) of edited copies of style transformations on legitimate inputs but not on adversarial inputs. 
Therefore, it is possible to identify a consistency threshold or train a classifier based on consistency to distinguish legitimate from adversarial instances. 
To formalize the process of our defense, we summarize notations in Table 1.
}
\vspace{-0.4cm}

\vspace{-0.4cm}
\begin{table*}[!htb]
	\setlength{\abovecaptionskip}{-0.05cm}
	\setlength{\belowcaptionskip}{-0.2cm}
\scriptsize
\centering
{\caption{\textcolor{black}{Notations used in this paper}}}
\begin{tabular}{ll}
\hline
{\color[HTML]{000000} Notation}                                            & {\color[HTML]{000000} Explanation}                                                                                   \\ \hline
{\color[HTML]{000000} $ \mathbb{U}$}                                       & {\color[HTML]{000000} the set of all instances in the sample space}                                                  \\
{\color[HTML]{000000} $ \mathbb{N}$}                                       & {\color[HTML]{000000} a manifold that consists of instances that act naturally w.r.t. a certain classification task} \\
{\color[HTML]{000000} $\mathbb{X} =\{x_1,x_2,\cdots ,x_n\}$, $x^*$}               & {\color[HTML]{000000} instances in $ \mathbb{U}$, adversarial example of x}                                                                       \\
{\color[HTML]{000000} C, $EC_x=\{ec_x^1,\cdots,ec_x^V\}$, $ cv_{x}^i \in \textbf{v}_x$}                                                   & {\color[HTML]{000000} classification function, edited copies for x, consistency value and vector for an instance x}                                                                       \\
{\color[HTML]{000000} E, Z, z}                                 & {\color[HTML]{000000} encoder $E$ maps input X into latent space Z, z is a latent vector}                                                                 \\
{\color[HTML]{000000} G, D, F}                                & {\color[HTML]{000000} Generator, Discriminator and Fully-Connected Layer for the pre-trained StyleGAN}                                                                \\
{\color[HTML]{000000} B, b}                                & {\color[HTML]{000000} B introduces noise inputs to generate stochastic details. $b$ is single-channel images consisting of Gaussian noise. }  
\\
{\color[HTML]{000000} AdaIn, $y_{sc,i},~y_{ba,i}$}                                 & {\color[HTML]{000000} adaptive instance  normalization with the scaled (sc) and biased (ba) parameters. }                                                                 \\
{\color[HTML]{000000} $w,~w',~w^*,~\hat{w^*}$}                        & {\color[HTML]{000000} intermediate latent codes $w$ via $F(z)$,encoded latent codes from E(x), fine-tuned w', shifted latent codes}                                                                            \\
{\color[HTML]{000000} $\rho,~\tau_{cv},~\tau_{v}$}     & {\color[HTML]{000000} false detecting rate, threshold for the consistency value, and threshold for evaluation of consistency vector}                                                                       \\
{\color[HTML]{000000}  $a_s$, $\gamma$ }                                          & {\color[HTML]{000000} latent style axis $a_s$ of the selected styles $s$, scalar used to specify the degree of transformation towards the style $s$ }                                                                        \\
\hline
\end{tabular}
\end{table*}
\section{Background and Related Work}
\subsection{Adversarial attacks}
Evasion attacks have long been studied on machine learning classifiers \cite{barreno2010security,lowd2005adversarial}, and are practical against many types of models \cite{biggio2013evasion}. These evasion attacks over neural networks are referred to as adversarial examples \cite{szegedy2013intriguing}. 
Let $ \mathbb{U}$ be the set of all instances in the sample space and $C$ for a classification model. 
Let $ \mathbb{N}$ be a manifold that consists of instances that act naturally with regard to a specific classification task, following a data generation process. $\mathbb{N}$ can be approximated by a set of legitimate instances for a classification task \cite{meng2017magnet}.
The goal of the adversarial example is to find a certain perturbation on $x$ to generate an adversarial example $x^* \in \mathbb{U}\setminus \mathbb{N}$ that fools a specific $C$ to misclassify, i.e. $ C(x^*) \neq C(x) $. 
\subsection{Adversarial defenses}
\textbf{\textit{Adversarial training.}} 
An intuitive way to build a robust classifier is to include adversarial information in the training process, which we refer to as adversarial training. For example, one may use a mixture of legitimate and adversarial examples in the training set for data augmentation \cite{szegedy2013intriguing}, or mix the adversarial objective with the classification objective as a regularizer \cite{goodfellow6572explaining}. Although this idea is promising, it is difficult to explain what attacks should be trained on, and how important an adversarial component should be. 
Besides, adversarial training involves retraining the protected model with adversarial examples, which may negatively impact the accuracy of clean inputs. Even a slight reduction in performance would result in significant losses for commercial or security-sensitive organizations. 
Furthermore, recent research works \cite{zhang2019limitations,song2018generative} have suggested that adversarial training can be defeated when the input image is generated by a generative model, especially for higher-dimensional datasets. 
In contrast to adversarial training, our detection with Clean-Only does not require adversarial information that would require intensive computing and storage resources to obtain.
\newline
\textbf{\textit{Detecting adversarial examples.}}  
\textcolor{black}{
The detection-based defense against adversarial examples aims to establish a detector $f_d: \mathbb{U} \rightarrow \mathbb{N} \cup \{J\}$. 
Here, $\mathbb{U}$ is the set of all instances for evaluation, $\mathbb{N}$ is the subset of $\mathbb{U}$ that consists of all clean instances, and $J$ is the judgment that the input is unlikely to be from the manifold of clean instances. 
The source of information for the judgment $J$ may be the clean instance dataset alone or in conjunction with extra information about the adversary, such as adversarial examples produced by some known attacks.
The methodology for the judgment $J$ could be a supervised approach to train separate classifiers or thresholds of specific evaluation metrics as detectors to recognize incoming adversarial examples. 
In contrast, supervised methods often require large amounts of computation, some of which result in a loss of accuracy on clean examples. 
The threshold-based detectors commonly apply transformations to the input and leverage the analysis of transformed and original inputs for detection, as detailed in the following section. 
}
\newline
\textbf{\textit{Transformation based detection defense.}} 
Generally, there are two kinds of transformation-based detection defense: detection based on the investigation for every single input and its single reconstruction copy, e.g., reconstruction error-based detection, and detection based on the investigation for every single input and its multiple transformation copies, e.g., consistency-based detection. For the reconstruction error-based detection, a large degree of protection against adversarial attack can be achieved by reconstructing inputs via generative models, such as Defense-GAN \cite{samangouei2018defense}, MagNet \cite{meng2017magnet}, FBGAN \cite{bao2018featurized}. 
The generative models, such as autoencoders, are conducted to detect the malicious input by evaluating the pixel-wise differences between the original input and its reconstructed version, and further purifying the malicious instance by moving it back towards the normal manifold distribution. 
Due to high computational complexity, these defense methods are limited in their ability to generate or reconstruct high-resolution images. Moreover, investigations of the pixel space of a single instance, such as those based on pixel-wise reconstruction errors, are insufficient to defeat powerful adversarial attacks that introduce minimal pixel perturbation to adversarial examples. 
The consistency-based detection approaches generally recognize the difference between the output of the DNN between real and adversarial images by performing certain spatial transformations on the input images, for example, \cite{ tian2018detecting,dathathri2018detecting}. It extends the investigation beyond a single instance perspective to the one encompassing its transformations. Transformations used in \cite{ tian2018detecting} include spatial rotation to train neural networks for detection with respect to varying rotation angles. Generally, existing approaches aim to train a classifier using both benign and adversarial images (derived from a specific attack A). Using rotated im ages as input, the trained detector (classifier) is used to identify the adversarial images from the attack A. 
However, such spatial transformations in pixel space could be targeted by the adversary and manipulated so as to bypass detection, such as spatially transformed adversarial examples \cite{xiao2018spatially}. 
In addition, training the target classifier with adversarial examples will consume considerable time and resources, particularly for such attack-specific detection.  
A neural fingerprinting (NFP) \cite{ dathathri2018detecting} based detection is achieved by verifying whether the prediction behavior of the model is consistent with a set of secret fingerprints. 
Once the secret fingerprints have been revealed to the adversary, the defense will be bypassed accordingly. Despite using less computing power, these methods may not work universally, and can only be applied to certain datasets. 
Unlike existing style transformation approaches, such as AdvMix \cite{gowal2020achieving}, our approach is characterized by its controllable nature.
AdvMix applies a heuristic mixing transformation to each pair of original and target images. The augmented copies during a transformation have no control on a specific attribute for AdvMix. As an example, during the transformation with respect to expression, the identification will also change. Our approach, however, is to first find clear correspondences between specific changes in style and changes in latent space, which can then be applied to any single instance instead of pair by pair. Our method allows us to perform a controllable transformation according to desired attributes with fine granularity. Besides, our defense could not be to introduce additional adversarial training and negatively affect the accuracy of protected models.

\subsection{Generative models for high-fidelity images} 
GAN has made tremendous progress in image synthesis in recent years.
A random latent code may be given to the GAN generator using adversarial learning, and a high-fidelity picture is created. 
The latent code may be translated into layer-wise codes using style-based generators such as StyleGAN/StyleGAN2, and then input into the convolution process through Adaptive Instance Normalization (AdaIN).
\textcolor{black}{
The non-linear mapping network $F$ initially generates intermediate latent codes $w$ from the normally distributed random noise vector $z$ in the latent space $Z$. For example, a sequence of completely linked layers is used to convert a 512-dimension $z$ to a 512-dimension $w$.
It is claimed that the $W$ space more appropriately reflects the learned distribution's disentangled character.
Following each convolution layer of the generator, $w \in W$ is converted to parameters specific to each channel via a separate affine transformation to manage adaptive instance normalization (AdaIN).
The AdaIN is defined as: 
\begin{equation}\footnotesize
AdaIN(x_i,y)=y_{sc,i}\frac{x_i-\mu(x_i)}{\sigma(x_i)}+y_{ba,i}
\end{equation}
Here, each of the feature maps x is normalized, and scaled $y_{sc,i}$, biased $y_{ba,i}$ independently using matching scalar components from the underlying style. 
StyleGAN2 contains a single style parameter per channel that modulates the convolution kernel weights to manage feature map variances. The tRGB blocks employ extra style parameters in the translation of feature map pictures to RGB images in addition to the style parameters.
}

\section{VASA-Defense Methodology}
\subsection{VASA-Defense overview}
\textcolor{black}{
Our VASA-Defense consists of three components: VASA model for latent representation and encoding (Section \ref{sec:vasa}), edited copy generation with style transformations via latent shifting and reconstruction using trained VASA (Section \ref{sec:shift}), and consistency based detection using edited copies (Section \ref{sec:detctor}), to address three key challenges:
(\textbf{C1}) \textit{How can a high-resolution image be reversely encoded into the latent representation of a given generator?}
(\textbf{C2}) \textit{
How can the latent codes be linked to styles, and how can edited copies be constructed with style transformations in a controlled manner?} 
(\textbf{C3}) \textit{Is it feasible to perform detection efficiently by utilizing style transformations? }
}
\newline
\textcolor{black}{
\textbf{Pipeline: }
First, a VASA model is developed to address challenge C1 in Section \ref{sec:vasa} \textit{ Disentangled latent representing and encoding}. It could be utilized to reverse encode high-resolution images into the latent space of a trained generative model with high reconstruction quality. 
Second, the controllable edited copy generation is conducted to address challenge C2 in Section \ref{sec:shift} \textit{Latent shifting and reconstruction based style transformations}. 
Based on a set of encoded latent codes and their associated style labels, we determine the style editing axis in latent space to identify correspondences between the changes in latent codes and the selected styles. 
To obtain a set of edited copies, we perform small shifts along the style editing axis of the encoded latent codes and reconstruct the shifted latent codes using the trained generator of VASA. 
Finally, we apply Section \ref{sec:detctor} \textit{Consistency based detection based on style transformations} to handle the challenge C3. 
Detection is based on the consistency of the predictions, derived from a pre-trained classifier, between an original image and its edited copies with style transformations. 
Details of these components are presented in the following sections. 
}
\subsection{Disentangled latent encoding}  
\label{sec:vasa}
\subsubsection{Scheme of the generative scheme VASA}
\textcolor{black}{
To address challenge C1, we propose a VASA model that maps high-resolution images to the latent codes of a powerful generative model with fine reconstruction quality. 
GANs typically comprise two neural networks, where a generator generates samples from random noise as input, and the samples are evaluated and differentiated using a discriminator. StyleGAN has demonstrated outstanding visual quality and fidelity for image generation with a learned intermediate latent space that precisely reflects the training data distribution. 
StyleGAN latent codes exhibit disentangled properties, which enables extensive manipulation of images when using a trained StyleGAN model  \cite{collins2020editing,harkonen2020ganspace,shen2020interpreting,wu2021stylespace,tov2021designing}.
}

\textcolor{black}{
StyleGAN also receives input from the random noise vector $z$ in the latent space $Z$. 
Accordingly, for image x' to be manipulated, we need to encode it into the trained generator's latent space to obtain its latent codes, i.e., image reverse encoding, which will be fed into the trained generator of the StyleGAN for the reconstruction of image x'. }
\textcolor{black}{
Therefore, an effective reverse encoding scheme is essential for image manipulation.
Specifically, we leverage the generative power of the state-of-the-art StyleGAN2 \cite{karras2019analyzing} and develop a reverse encoding method utilizing adversarial learning to construct an autoencoder that takes full advantage of the editing capabilities of StyleGAN2 to conduct controllable and realistic manipulations of a given image. VASA's encoder can encode the image to the latent representation of StyleGAN2. Then, the generator serves as a decoder to reconstruct the modified latent codes. The overall model is depicted in Fig. \ref{fig:scheme}(a). 
During the training of the VASA model, we add the additional encoder $E$ to map given images to the latent space of a generative model $G$ with fine reconstruction quality via the decoder $D$. $z$ represents the random noise vector in the latent space $Z$, which is the input to the function $F$. $F$ could be considered as part of the generator, adopted to map $z$ to the W latent space to reflect the learned distribution's disentangled nature better, as the feature extractor of G. The discriminator D aims to distinguish inputs $(x,w')$ and $(G(w,b),w)$, providing the information for gradient updates. The details of training and implementation of VASA are in Sections 4.2.2-4.2.3. After training the VASA, we conduct the style transformations as the inference stage. Only the trained generator G and encoder E of the trained VASA are used during inference. E maps a target image to a latent vector in the latent space of the trained G. Afterwards, each level's feature maps are further enhanced by adding a suitable random noise map $b' $ to recover additional stochastic details. Reconstruction is conducted via G(w',b'). Details are in Section 4.3. 
}

\subsubsection{Implementation of the VASA}
\textcolor{black}{
As it is not stable to update all components of VASA at the same time, we train the VASA in an alternating update manner (adversarial learning) and use a smooth version of the rectifier activation function SoftPlus \cite{glorot2011deep}, defined as $f(\cdot) = softplus(\cdot) = log(1 + exp(\cdot))$ in the loss function. 
We treat F as a deterministic map from the pre-trained StyleGAN2. Therefore, the loss function follows the BiGAN \cite{donahue2016adversarial}, given as follows:}
\begin{equation}\footnotesize
\begin{aligned}
\underset{G,E}{max}\underset{D}{max}\mathbb{E}_{x\sim p_x}[\mathbb{E}_{w\sim p_{E(\cdot|x)}}[log(1-D(x,w))]]\\+\mathbb{E}_{w\sim p_w}[\mathbb{E}_{x\sim p_{G(\cdot|w)}}[log~D(x,w)]] 
\end{aligned}
\end{equation}
\textcolor{black}{
The training of VASA uses alternating updates to optimize the loss function. In particular, we partially update some components while leaving others unchanged.
Step I, the discriminator D and encoder E are updated with the fixed G and F. 
Step II, we update the generator G with others fixed. 
Step III involves updating the latent space autoencoder (i.e., networks G and E) with others fixed. 
For updating the weights, we use the Adam optimizer with $\beta_1 = 0.001$ and $\beta_2 = 0.99$. For non-growing architectures (i.e., MLPs of D and F), we use a learning rate of $0.002$, and a batch size of 128. In growing architectures (i.e., G and E), the learning rate and batch size are dependent on the resolution.
Typically, the main loss function and regularization terms are optimized simultaneously. In the VASA training, we use lazy regularization, which means that the regularization terms can be computed less frequently than the main loss function, thereby significantly reducing the computational cost and the memory usage. As a result, regularization is performed only once every 16 mini-batches.
}
\subsubsection{Architecture of Components}
\label{sec:comp}
\textcolor{black}{
Fig. 3 (a) illustrates the architecture of components for VASA. 
Latent space $w$ has the same role as the intermediate latent space in StyleGAN2. 
F is used to map $z$ to latent space $w$ for feasible disentanglement.  
The generator G utilizes 8 levels of resolution, i.e., from $4 \times 4$, $8 \times 8$ to $1024 \times 1024$, to reveal different levels of abstraction of style information, used to learn spatially invariant style $y=(y_{sc,i},y_{ba,i})$. G consists of 18 layers of MLP, two layers for each level of resolution. 
For instance, the first four layers (4 and 8 resolutions) reveal the facial shape or pose, and the middle four layers (16 and 32 resolutions) describe the facial details, e.g., eyes, nose, mouth details, and expression. The rest layers illustrate color and other detailed textures. 
}

\textcolor{black}{
The affine transformation A is used to specialize the style information $w$ at each layer, namely scaled $y_{sc,i}$ and biased $y_{ba,i}$ scalar components that control adaptive instance normalization (AdaIN, i.e., modulation operation, Mod) on feature map $x_i$ after normalizing each convolution layer. 
Namely, A learns the spatially invariant styles, scaled $y_{sc,i}$ and biased $y_{ba,i}$ scalar components from $w$, that are then applied to the normalization of every layer of G.  
The function B is used to introduce noise inputs to generate more stochastic details. 
The set of $b$ is single-channel images consisting of uncorrelated Gaussian noise, fed into the feature map of each layer of G. The noise image is broadcasted to all feature maps using learned per-feature scaling factors (similar to AdaIN) and then added to the output of the corresponding convolution.
To improve the G, similarly to \cite{karras2019analyzing}, the set of styles that are input to the revisited Instance Normalization layers in G are related linearly to the latent variable $w$. The normalization and modulation are operated on the standard deviation alone. Besides, bias and noise operations are moved outside the style block, where they operate on normalized data. 
Skip connections are adopted instead of progressive growth. Specifically, the contributions of RGB outputs are unsampled and summed, corresponding to different resolutions. 
In the discriminator, the downsampled image is provided to each resolution block of the discriminator. 
Bilinear filtering is used in all up and downsampling operations. 
We start with low-resolution images ($4\times 4 $ pixels) and progressively increase the resolution by smoothly blending in new blocks to E and G. For the F and D networks, we implement them using MLPs. The Z and W spaces and all F and D layers have the same dimensionality in all our experiments. F has eight layers, and we set D to have three layers. 
For high-resolution images, e.g. $1024\times 1024$ face images, the latent codes $z$ and $w$ are $512$-dimensional feature vectors. }

\begin{figure*}[!htb]
	\centering
	\setlength{\abovecaptionskip}{-0.05cm}
	\setlength{\belowcaptionskip}{-0.2cm}
	\subfigure[Scheme of VASA.]{
		\includegraphics[width=2.5in]{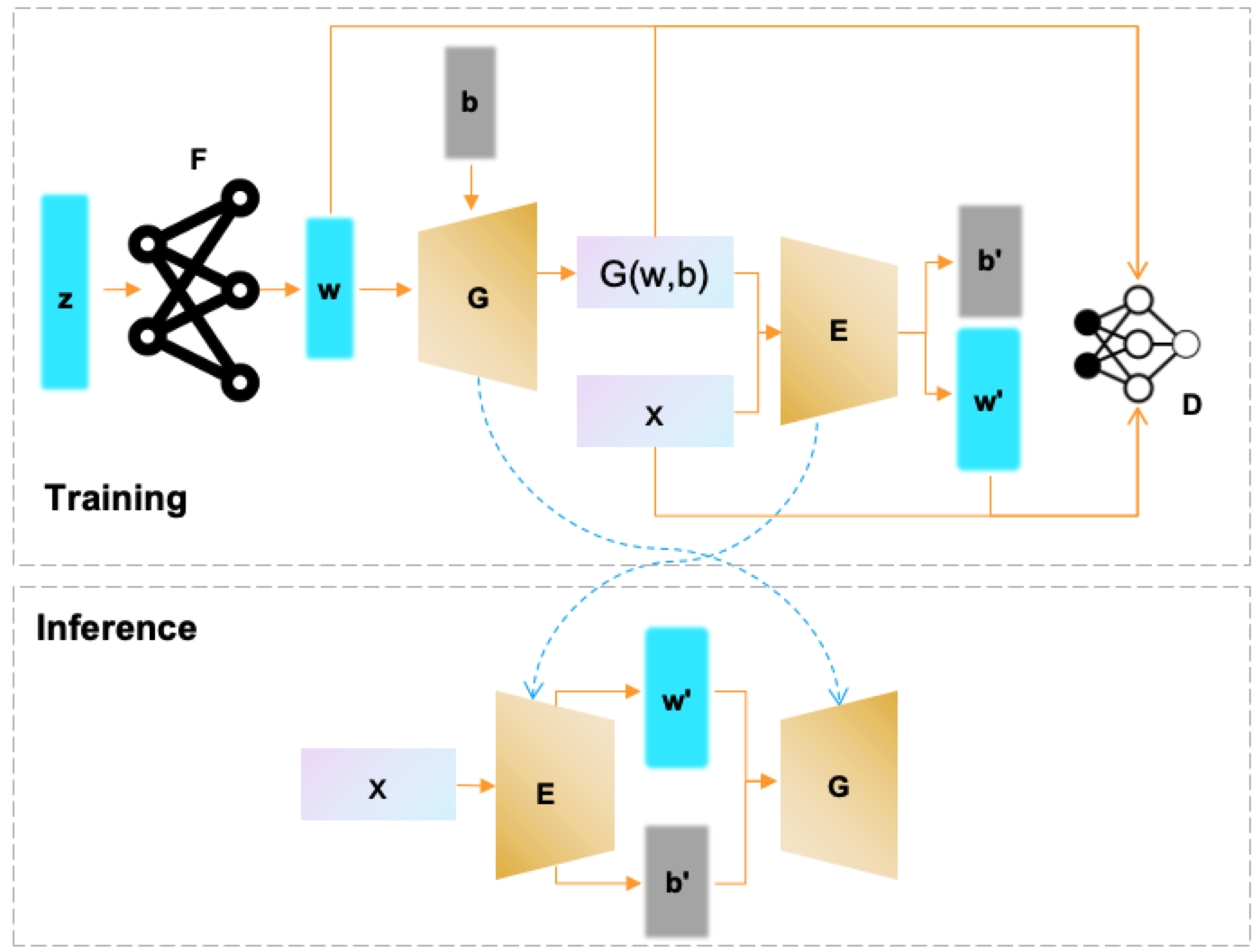} }
	\subfigure[Illustration of components.]{    \includegraphics[width=4.2in]{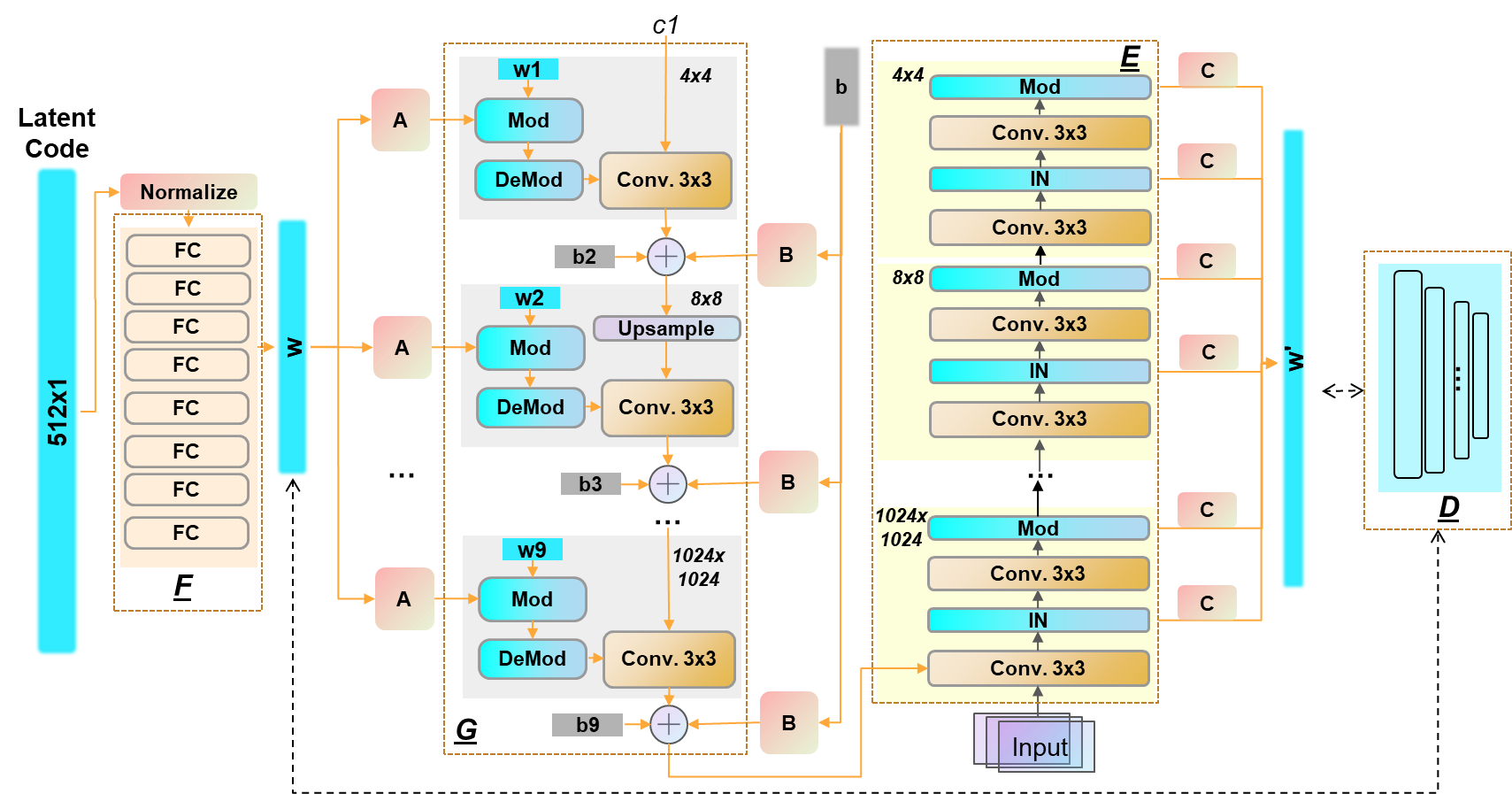}     }
	\caption{Illustration of the VASA. \textcolor{black}{ (a) Structure of VASA for training and inference. 
	\textit{\scriptsize{During training of the VASA model, additional encoder $E$ is added to map given images to latent space of a generator $G$. D and F are used to improve generative ability and provide gradient information to update the entire generative model. After training the VASA, only the trained generator G and encoder E of the trained VASA are used for style transformations.}} (b) Architecture of component F, G, E and D of VASA. \textit{\scriptsize{Details are in Section \ref{sec:comp}.} }}} 
	\label{fig:scheme}
\end{figure*}

\textcolor{black}{
The architecture of encoder E is as illustrated in Fig. \ref{fig:scheme} (b). As G is driven by a style input at every layer, E is designed symmetrically, so that style information of the encoded image can be extracted from a corresponding layer. Instance Normalization (IN) layers are used to provide instance averages and standard deviations for every channel as extracted style information of the encoded image, similar to \cite{pidhorskyi2020adversarial}.
If $y^E_i$ is the output of the i-th layer of E, the IN mod style at that level provides instance averages and standard deviation statistics, $ \mu(y^E_i )$ and $\sigma(y^E_i)$, revealing styles at every channel. The IN module also provides the normalized version of the input as output, which continues down the pipeline with no more style information from that level. The output styles from the encoder are combined before mapping them onto the latent space, via the following multilinear map. 
Here, $C_i$ is the learnable parameter, and N is the amount of the layer.
\begin{equation}\footnotesize
w=\sum^{N}_{i=1}C_i\begin{bmatrix}
\mu(y^E_i )\\ 
\sigma(y^E_i)
\end{bmatrix}
\end{equation}
}
\subsection{Latent shifting and reconstruction based style transformations}
\label{sec:shift}
\textcolor{black}{
After the training of VASA, the controllable edited copy generation can be conducted as the inference stage shown in Fig. \ref{fig:scheme} (a), providing the trained G and E. In general, the generation of edited copies based on latent shifting and reconstruction involves three procedures: (1) encoded latent code fine-tuning and noise embedding; (2) latent and style mapping; and (3) reconstruction via the trained generator G. 
Specifically, given an underlying image $x$, we first use the provided encoder E of the trained VASA to map a target image $x$ to the latent codes via $E(x)=w'$. The encoded image may not be from the training data for the trained generator G. Therefore, the $w'$ is fine-tuned to $w^*$ derived from the original generative space of the trained generator G via the \textit{Encoded latent code fine-tuning} (Section \ref{sec:invert}). Simultaneously, we aim to find a suitable random noise map $b' $ added to each level feature map to recover more stochastic level details.  
A given latent code vector represents a fixed image with fixed features within the latent space. If this vector is shifted across an axis, the corresponding semantic features of the image are altered. As long as the vector is only modified in the direction of a specific feature in latent space, everything remains the same except for the feature derived from which the vector (latent codes) is being shifted. 
Consequently, we find the style editing direction (axis) $a_s$ in the latent space to reveal correspondences between the change of latent codes and some selected non-essential styles (detailed in Section \ref{sec:correspondence} Latent and style mapping). 
Given the style editing direction vector $a_s$, we manipulate the style $s$ via manipulating the encoded latent codes $w*$ of the given image $x$ as follows: 
\begin{equation}\footnotesize
    \hat{w^*}=w^*+\gamma a_s
    \label{eq:md}
\end{equation}
Here, $\gamma$ is the scalar used to specify the degree of transformation towards the style $s$. 
Then, we obtain a set of edited copies by conducting $m$ times small shifts ($\gamma_i,~i=[1,\cdots,m]$) on the encoded latent codes $w^*$ along the latent style axis $a_s$ of the selected styles $s$, following by reconstructing the shifted latent codes $\hat{w^*}=\{\hat{w_1^*},\hat{w^*_2},\cdots,\hat{w^*_m}\}$ to a set of transformed images by passing through the trained generator G via $G(\hat{w^*_i},b'_i)$.
}
\subsubsection{Encoded latent code fine-tuning}
\label{sec:invert}
\textcolor{black}{
Without the encoder, the searching range for each latent code fine-tuning would be the entire latent space, resulting in extremely large computing overheads. The search range is limited to a smaller area around the encoding vector $E(x)$ by using the encoder E. 
Furthermore, the encoder is trained using dataset-wise optimization to approximate global generalization. Therefore, the encoding embedding for a specific instance has a deviation. The embedding ability of the encoder is not perfect enough to represent each instance in the latent space, especially for these unseen instances, due to the generalization deviation. Therefore, we apply instance-specific latent code fine-tuning to refine the encoding mapping deviation. 
}

The trainable parameters for the fine-tuning are the components of encoded latent codes, started at $w_0=E(x)=w'$, as well as components in all noise maps $b'_i$ initialized as $N(0,I)$. 
Latent fine-tuning searches for an optimized vector $w^*$ that minimizes the loss function that measures the similarity between the given image and the image generated using $w^*$ and $b'_i$. 
The next step is to design per-layer noise maps, denoted by $b'_i \in R^{r_i \times r_i}$, where $i$ is the layer index and $r_i$ denotes the resolution of the $i^{th}$ noise map. 
The generator/encoder in $1024\times 1024$ resolution has 17 noise inputs, i.e., two for each resolution from $8\times 8$ to $1024\times 1024$ pixels and one more noise input added after the learned $4 \times 4 $ pixels. 
We compute $\mu_w =  \mathbb{E}_x E(x)$ by encoding 10,000 random clean inputs through the mapping network encoder at first. 
The scale of $w$ is approximated by computing $\sigma_w^2 =  \mathbb{E}_x ||E(x) -\mu_w||_2^2$ , i.e., the average square Euclidean distance to the center. 
Since the computational graph for $G(w)$ is given, $w^*$ and noise maps can be calculated via gradient descent methods, taking the gradient of G with regard to w and the noise maps. 
To evaluate the similarity between the input image and the generated image, and the regulation term of noise maps during optimization, a loss function is a weighted combination of the LPIPS (Learned Perceptual Image Patch Similarity) \cite{zhang2018unreasonable} distances loss, Mean Squared Error (MSE) loss and regularization term:
\begin{equation}\footnotesize
L= \lambda_P L_{lpips} (G(w^*),x) + \lambda_M ||G(w^*)-x||^2_2+ \alpha \sum_{i,j} L_{i,j} \end{equation}
$\lambda_P=\lambda_M=\alpha = 10^5$ in all tests, and N is the number of samples.  
The image quality term is the LPIPS distance between the target image $x$ and the synthesized image. For increased performance and stability, we downsample both images to $256\times 256$ resolution before computing the LPIPS distance. 
Regularization of the noise maps is performed on multiple resolution scales. Therefore, for each noise map greater than $8\times 8$ in size, we form a pyramid down to $8\times 8$ resolution by averaging $2\times 2$ pixel neighborhoods and multiplying by 2 at each step to retain the expected unit variance. These downsampled noise maps are used for regularization only and have no part in synthesis. 

The regularization term for noise map $n_{i,j}$ is then where the noise map is considered to wrap at the edges. 
The regularization term $L_{i,j}$ for the noise map $n_{i,j}$  is thus the sum of the squares of the resolution-normalized auto-correlation coefficients at one pixel shifts horizontally and vertically, which should be zero for a normally distributed signal. Let $r_{i,j}$ be the resolution of an original $(j = 0)$ or downsampled $(j > 0)$ noise map so that $r_{i,j+1}=r_{i,j}/2$ \cite{karras2019analyzing}.
\begin{equation}\footnotesize
\begin{aligned}
L_{i,j}=(\frac{1}{r^2_{i,j}}\sum_{x,y}n_{i,j}(x,y)n_{i,j}(x-1,y))^2\\+(\frac{1}{r^2_{i,j}}\sum_{x,y}n_{i,j}(x,y)n_{i,j}(x,y-1))^2
\end{aligned}
\end{equation}
x and y can be the same image. 
After each optimization step, we re-normalize all noise maps to zero mean and unit variance. The optimization is run for 700 iterations using the Adam optimizer with default parameters. The maximum learning rate is $\lambda_{max} = 0.1$, and it is ramped up from zero linearly during the first 50 iterations and ramped down to zero using a cosine schedule during the last 250 iterations. In the first half of the optimization, Gaussian noise is added to $w$ when evaluating the loss function as $\hat{w} = w + N(0, 0.05 \times \sigma_w t^2)$, where t goes from one to zero during the first half of the iterations. Such noise could add stochasticity to the optimization process. 
\subsubsection{Latent and style mapping}
\label{sec:correspondence}
\textcolor{black}{
In order to facilitate style transformations, it is necessary to identify the style axis in the latent space in order to determine correspondences between the change of a specific style and the change of latent space, i.e., latent and style mappings. 
In general, logistic regression is used to determine a separate hyper-plane in the latent space that separates the features of interest (style) from other features in a supervised manner, using the labeled latent codes grouped according to styles. 
It is feasible to obtain a well-labeled image dataset from the public image cognitive service APIs, e.g., the labeled face image recognized from the Azure Microsoft (https://azure.microsoft.com/en-us/services/cognitive-services/face/). There are abundant labels available, including rotation angle, expression, glasses, etc. One can select one non-essential style $s$ as a feature of interest, such as facial expression, and create a face image group $G_s$ that carries this label. This allows us to investigate the change in latent codes due to changing selected styles (e.g., from sad to happy) within the group. 
}
\textcolor{black}{
The style axis in the latent space to reveal correspondence can be described using a direction vector \textbf{$a_s$}, with the same dimensions as latent codes $w$. The basic method to get \textbf{$a_s$} is as follows:
Taking two samples $p$ and $q$ at each time from the $G_s$, the approximate $\vec{a_s}$ is the ratio via dividing the difference of their latent codes $w^*$ by the difference of their labels, and the final \textbf{$a_s$} can be considered as the average $\vec{a_s}$ on multiple sampling iterations. To further improve the efficiency of the inference, we can take the median of the labels as the dividing line. Labels lower than this value are changed to 0, and labels higher than this value are changed to 1. By applying logistic regression to solve the objective function of a binary classification problem, $a_s \times w^*+b$. Here, $w^*$ is the encoded and fine-tuned latent code vector, and the obtained $a_s=\{a_1^s,\cdots,a^s_{|w^*|}\}$ can be approximated as the style direction vector \textbf{$a_s$}. 
Finally, the edited copies can be obtained by feeding the generator with shifted latent vectors by applying $\hat{w^*}=w^*+\gamma a_s$ (Eq. \ref{eq:md}) using the learned $a_s$ and different scale factors (i.e., $\gamma$ as the magnitude of change). 
}
\subsection{Consistency-based detection using style transformations}
\label{sec:detctor}
\textcolor{black}{
In this section, we use the consistency of classification between an original image and its edited copies with non-essential style transformations as an indicator of detection. 
We perform performance analyses by varying the amount of information available to learn the detector and its implementation settings. 
First, we develop \textit{Detection with Clean-Only Consistency} in Section \ref{sec:cleanonly} where the only information used in training the detector is some clean data against adversaries with black-box knowledge about the detector.
Additionally, we explore the \textit{Detection with Adversarial Supervision} in Section \ref{sec:supervise} where the information used in training the detector is some clean data and adversarial examples counterparts derived from specific attacks, then test the performance on other unseen attacks. Then, we extend the \textit{Detection against White-Box Attack} in Section \ref{sec:white} to enhance the detection mechanism against an adversary with white-box knowledge about classifier+detector, i.e., defending adaptive attacks. Details of the adversarial settings are given in Section \ref{sec:attacksetting}.
}
\subsubsection{Detection with Clean-Only consistency}
\label{sec:cleanonly}
\textcolor{black}{
For the purposes of this scenario, we assume that the inputs to the detection mechanism are only a set of clean data used to train the victim model.
Using the edited copy with non-essential style transformations, we will find a detector that rejects instances that have malicious behaviors. 
The detection indicator should easily differentiate legitimate and adversarial instances, using clean instances only. In addition, it should be attack-independent. 
Rather than using the element-wise reconstruction error, we use the consistency of classification between the original image and the edited copy, as determined by a pre-trained classifier (which may or may not be the same as the targeted classifier), as an indicator for the detection.
}

\textcolor{black}{
Given an investigated image $x$ and its edited copies $EC_x=\{ec_x^1,\cdots,ec_x^V\}$ via style transformations, we use a pre-trained classifier $C$ to record the predictions for both of them. 
Classification-based consistency is defined as the ratio of unchanged classification predictions of these edited copies compared to those of the original image:
\begin{equation}\footnotesize
    Consistency=\frac{|\{ec_x^i:C(x)=C(ec_x^i)\}_{i=1}^V|}{V}
    \label{eq:consistency}
\end{equation}
We find the averaged consistency of a clean instance is significantly higher (almost 100\% consistent) than that of an adversarial one (below 20\%), as shown in Fig. \ref{fig:illustrate} (b) on MNIST and Fig. \ref{fig:threshold} on high-resolution face images from FFHQ. 
Then, we can calculate the consistency values for a given clean dataset $DL $ and sort the consistency values in descending order. 
Therefore, it will be possible to use a consistency threshold derived solely from clean examples, as a detector, to determine whether an input is clean or suspicious based on its consistency value. 
Specifically, the threshold will be decided as the consistency value at the top $\rho\%$ percentile of the sorted consistency values for $DL$. As $\rho\%$ increases, a smaller consistency value for $DL$ will be applied as a threshold.
Namely, the detector could correctly distinguish $\rho\%$ instances in the clean dataset $DL$ as clean (consistence value $>$ threshold) and mistakenly reject no more than $1-\rho\%$ clean instances.
 }

\textcolor{black}{ \textbf{Pipeline}: 
\textit{Step I.} Given a set of clean images $DL=\{x_1,\cdots,x_N\}$ are used to produce the detector, we generate a set of edited copies for each image. Specifically, an image is first mapped to the latent code $w'$ by the encoder and then fine-tuned to $w^*$. Next, $V$ times latent shifting operations are iteratively applied to each clean image $x$ along a specific latent style axis in order to generate a set of $V$ edited copies $EC_x=\{ec_x^1,\cdots,ec_x^V\}$. 
\textit{Step II.} For each image $x \in DL$, we feed its edited copy set $EC_x$ into a pre-trained classifier $C$ and record the predictions for both $x$ and all $ec_x^i \in EC_x$. 
We then calculate the consistency value $cv_x$ for each $x \in DL$ according to Eq. \ref{eq:consistency}, as the indicator for detection.
\textit{Step III.} The threshold $\tau_{cv}$ will be decided at the top $\rho\%$ percentiles of consistency values of $DL$, sorted in descending order. 
\textit{Step IV.} During inference, a given $x$ will be recognized as malicious and rejected when $cv_x$ is less than the specified consistency threshold $\tau_{cv}$, otherwise as legitimate. 
}
\subsubsection{Detection with adversarial supervision}
\label{sec:supervise}
\textcolor{black}{In this scenario, we assume that the inputs for the detection mechanism are a set of clean data used to train the victim model and the adversarial examples derived from specific adversarial attacks conducted against part of the clean dataset. The detector can be decided using supervised learning techniques to distinguish adversarial examples from legitimate images. The attacker is assumed to be unaware of the detection mechanism.} 
Let the victim classifier $C$ be a classification task with $K (K \geq 2)$ categories, on a clean set $\mathcal{DL} = \{x_i \}^N_{i=1}$ associated with labels $\{t_i\}^N_{i=1}, t_i \in \{1,\cdots,K\}$. Based on labels and predictions, the dataset produces partitions $\mathcal{DL}=\bigcup \{\mathcal{DL}_k=\{x:t_x=k\}\}$ and $\mathcal{DL}^f =\bigcup \{\mathcal{DL}_k^f=\{x:C(x)=k\}\}$. 
The Lp norm bounded adversarial examples set is defined as $\mathcal{DL}^{\prime}=\{x+\delta: f(x+\delta) \neq t, C(x)=t, x \in \mathcal{DL}, \delta \in \mathcal{S}\}, \mathcal{S}=\left\{\delta \in \mathbb{R}^{d} \mid\|\delta\|_{p} \leq \epsilon\right\}$.
\textit{Detection with Adversarial Supervision} aims to build binary classifiers as detectors, $H=\{h_k\}^K_{k=1}$. Here, the objective for $h_k$ is to discriminate whether a sample classified as label $k$ is clean sample or an adversarial sample. Specifically, given an input sample x, the initial step is to determine the predicted class category of x from $k=C(x)$, followed by obtaining its m edited copies $EC_x=\{ec_x^i\}_{i=1}^m$ and associated class labels from $C(\cdot)$ to construct the consistency vector $\textbf{v}_x=\{v_1,\cdots,v_m\}$, where $v_i=1~if~C(ec_x^i)=k~otherwise~0$. 
Next, with the k-th detector, we determine x is a natural example when $h_k(\textbf{v}_x)=1$, otherwise an adversarial one. 
Since minimizing the classification error corresponds to minimizing that of individual detectors \cite{yin2019adversarial}, each k-th detector with parameter $\theta_k$ is then trained on:
\begin{equation}\footnotesize
\begin{aligned}
 \theta_{k}^{*}=\underset{\theta_{k}}{\arg \min } \mathbb{E}_{x \sim \mathcal{DL}^{\prime}{ }_{k}^{f}}\left[L\left(h_{k}^{\theta_{k}}\left(\textbf{v}_x\right), 0\right)\right]\\+\mathbb{E}_{x \sim \mathcal{DL}_{k}^{f}}\left[L\left(h_{k}^{\theta_{k}}\left(\textbf{v}_x \right), 1\right)\right]
\end{aligned}
\end{equation}
The loss L is used to measure the distance between the true label and the output of $h_k$, such as the binary cross-entropy loss. 
\subsubsection{Detection against White-Box adaptive attack}
\label{sec:white}
\textcolor{black}{
In this scenario, we assume adaptive attacks that are aware of the detector and adapt their attack strategies accordingly, e.g., the classifier and detector targeted attack settings in Section \ref{sec:attacksetting}. 
We also assume the clean dataset and their adversarial counterparts are available to train the detection mechanism. 
We incorporate attacks with white-box knowledge about both classifier and detector into the training objective by decreasing the upper bound by applying the unconstrained objective \cite{madry2017towards,yin2019adversarial}: 
\begin{equation}\footnotesize
\begin{aligned}
    \min _{\theta_{k}} \mathbb{E}_{x \sim \mathcal{D}_{\backslash k}}\left[\max _{\delta \in \mathcal{S}} L\left(h_{k}^{\theta_{k}}\left(\textbf{v}_{x+\delta } \right), 0\right)\right]\\+ \mathbb{E}_{x \sim \mathcal{D}_{k}}\left[L\left(h_{k}^{\theta_{k}}\left(\textbf{v}_x \right), 1\right)\right]
    \end{aligned}
\end{equation}
The detectors are learned with in-class natural instances and adversarial instances generated from samples outside the class. To solve the inner maximization, we employ an iterative PGD approach. Regular SGD optimizers are used to solve the outer minimization. 
Therefore, if L is expressed in binary cross-entropy terms, then the objective optimization is also a GAN-like problem, and the convergence law of GANs is straightforward to prove for the out-of-class distribution.
Similar to \cite{yin2019adversarial}, the Gibbs distribution $p(\textbf{v}_x, k)=\frac{\exp \left(-E_{\theta_{k}}(\textbf{v}_x)\right)}{Z_{\Theta}}$ could be used to calculate the joint probability of an instance and a label within energy-based learning framework. $E_{\theta_{k}}(\textbf{v}_x)=-z\left(h_{k}(\textbf{v}_x)\right)$, and $Z_{\Theta}=\sum_{k} \int \exp \left(-E_{\theta_{k}}(\textbf{v}_x)\right) d x$ is represented here by the partition function, which is a normalizing constant. The Bayes classification rule can therefore be used to create a generative classifier $H(\textbf{v}_x)=\arg \max _{k} p(\textbf{v}_x, k)=\arg \max_{k} z\left(h_{k}(\textbf{v}_x)\right)$. In addition, inputs with low probability can also be rejected using $p(\textbf{v}_x, k)$. A reject option is implemented by thresholding the logit output of the k-th detector, with k as the estimated label. Detection of malicious input is based on rejecting samples.
The detection proceeds as follows: given an input $x$, its consistency vector $\textbf{v}_x$, its estimated label from $k = C(x)$,  $x$ will be recognized as malicious and rejected when $z(h_k(\textbf{v}_x))$ less than a given threshold $\tau_{v}$, otherwise considered as legitimate and classified as $k$.
Note that, for these two detectors with supervision, the input for the binary classifier is the low-dimensional consistency vector $\textbf{v}_x$ instead of high-dimensional samples in the pixel space.
}
\section{Experiments}
\subsection{Data and setups}
\subsubsection{Datasets}
The performance of VASA-Defense is evaluated against the state-of-the-art adversarial attacks on three datasets: MNIST \cite{lecun1998gradient}, LSUN car (resized to 512 $\times$ 512) and cat (256 $\times$ 256) \cite{yu2015lsun}, and FFHQ \cite{karras2019style,karras2019analyzing} (Face). Flickr-Faces-HQ (FFHQ) contains 70,000 high-quality images at $1024\times 1024$ resolution. 
Our current focus is on single-domain data, i.e., data that pertains to a specific domain, such as human faces, cats as animals, or cars as vehicles.
We will then move on to multi-domain data in the future step.
The bottleneck is the data scale and the computation sources for the generative model on multiple domains, which is still an open question among the generative community. 
We randomly select 20,000 clean examples to train the VASA model for each dataset. We randomly select 5,000 clean images (named CLE, labeled 0) and generate 5,000 successfully attacked adversarial examples (named ADV, labeled 1), respectively. These datasets are used to test the efficiency of VASA-Defense. A further 2,000 clean instances are chosen at random as the validation data (named VAL) to determine thresholds or train the detector discriminator. 
\textcolor{black}{
The victim classifier for MNIST using the setting in \cite{carlini2017towards} with an accuracy of  $99.4\%$. For the LSUN and FFHQ, we train a classifier using the setting in \cite{simonyan2014very} with an accuracy of $98.2\%$ and $ 94.7 \%$ for identification recognition. 
We use the fast gradient sign method (e.g., FGSM \cite{kurakin2016adversarial}), DeepFool ($L^{\infty}$) \cite{moosavi2016deepfool,carlini2017towards}, $L^{\infty}$ Projected Gradient Descent Attack \cite{madry2017towards}, Basic Iterative Method that minimizes the $L^{2}$ and  $L^{\infty}$ distances \cite{kurakin2016adversarial}, Momentum Iterative Method attack 
\cite{dong2018boosting} and C\&W ($L^{2}$) attack \cite{moosavi2016deepfool,carlini2017towards} for the experiments, implemented using Fooling box \cite{rauber2017foolbox}. The detailed hyperparameters are given in Appendix \ref{sec:hyper}. 
}
We generate 1000 edited copies by conducting latent shifting 1000 times along a specific style axis in the latent space. The edited copies are fed into a pre-trained classifier, such as the VGG16 neural network used in our test, to calculate the classification consistency. 
\subsubsection{Cross-attack evaluation}
\textcolor{black}{
For assessment of VASA's performance in cross-attack scenarios, we train the detection model on one set of attacks and then record the defensive performance against the other unseen attacks. For example, we train the defense using the FGSM attack, and evaluate the performance against other unseen adversarial attacks. The detailed results and analysis are given in Section \ref{sec:crossevaluation}. }
\textcolor{black}{\subsubsection{Attack Settings}\label{sec:attacksetting}
We assume that the adversaries have different knowledge about the victim classifier and detector. 
To evaluate our defense against adversaries with different knowledge, we consider three attacking scenarios in terms of the knowledge about the victim classifier and the detector defense. }

\textcolor{black}{\textbf{(A-I) Classifier targeted attack. }
An adversary's goal of this attack is only to mislead the unsecured victim classifier model.
The detection is viewed as a post-processing step for dealing with malicious inputs. Adversaries are unaware of detection techniques, which is an assumption that is commonly employed by current detection-based defenses. They only focus on generating adversarial examples to maximize the prediction errors on the victim classifier. An adversarial sample is calculated via minimizing the following loss:
\begin{equation}\footnotesize
    L\left(x^{\prime}\right)=S\left(C\left(x^{\prime}\right)\right)_{y}-\max _{i \neq t} S\left(C\left(x^{\prime}\right)\right)_{i}
    \label{eq:black}
\end{equation}
Here, $S\left(C\left(x^{\prime}\right)\right)$ is the logit output. It is a method of untargeted attack based on the C\&W attack.} 

\textcolor{black}{\textbf{(A-II) Detector targeted attack. }
We assume that an adversary will produce adversarial instances by defeating only the detectors, based on the obtained consistency vectors of a given input. To construct the single detection function H, the logit output of i-th detector is used as the i-th output of H: $D(H(\textbf{v}_x))_i = D(h_i(\textbf{v}_x))$. The perturbed sample $x'$ is calculated on the following loss: 
\begin{equation}\footnotesize
L\left(x^{\prime}\right)=-\max _{i \neq t} D\left(H\left(\textbf{v}_{x^{\prime}}\right)\right)_{i}
\label{eq:gray}
\end{equation}
As the detector is a binary classifier, whose logit of output is a single value, we apply a detection rule for adversarial examples by applying a low logit value to indicate an adversarial example. Therefore, minimizing the negative logit is equivalent to reducing the perturbed example's likelihood of detection. 
}

\textcolor{black}{
\textbf{(A-III) Detector + Classifier targeted attack (adaptive attack).} 
In this case, the adversary will produce perturbed samples to bypass the detection system while at the same time fooling the victim classifier. Additionally, we assume the adversary could obtain the consistency vectors of a given input and feedback regarding whether the detector recognizes the modified input. When realizing the combined attack, we consider both loss functions for the detector and classifier. 
The implementation of combining these two loss functions is to optimize x' for the adversarial goal if x' is not currently an adversarial instance on the classifier. Otherwise, x' is optimized for fooling the aggregated detector.
The combination loss is formally given as follows:
\begin{equation}\footnotesize
\begin{aligned}
L\left(x^{\prime}\right)= \begin{cases}z\left(C\left(x^{\prime}\right)\right)_{t}-\max _{i \neq t} S\left(C\left(x^{\prime}\right)\right)_{i} \\ \text { if } S\left(C\left(x^{\prime}\right)\right)_{t} \geq \max _{i \neq t} S\left(C\left(x^{\prime}\right)\right)_{i}, \\ -\max _{i \neq t} S\left(H\left(\textbf{v}_{x^{\prime}}\right)\right)_{i}  else. \end{cases}
\end{aligned}
\label{eq:white}
\end{equation}
}

\textcolor{black}{
We summarize the attack setting and according strategies of detector in Table 2. 
\begin{table}[!htb]
	\setlength{\abovecaptionskip}{-0.05cm}
	\setlength{\belowcaptionskip}{-0.2cm}
\scriptsize
\centering
\caption{Attack setting and according strategies of detector}
\begin{tabular}{l|ll|l}
\hline
\multirow{2}{*}{Attack} & \multicolumn{2}{l|}{Adversary Knowledge}    & \multirow{2}{*}{Detection Strategy}                                                                             \\ \cline{2-3}
                        & \multicolumn{1}{l|}{Classifier} & Detector  &                                                                                                                 \\ \hline
(A-I)                   & \multicolumn{1}{l|}{White-box}  & Black-box & 4.4.1 Clean-Only                                                                                                \\ \hline
(A-II)                  & \multicolumn{1}{l|}{Black-box}  & White-box & \begin{tabular}[c]{@{}l@{}}4.4.2 Adversarial Supervision  and\\ 4.4.3 Detection against White-Box\end{tabular}  \\ \hline
(A-III)                 & \multicolumn{1}{l|}{White-box}  & White-box & \begin{tabular}[c]{@{}l@{}}4.4.2 Adversarial Supervision  and \\ 4.4.3 Detection against White-Box\end{tabular} \\ \hline
\end{tabular}
\end{table}
}
\subsubsection{Complexity of the once-off model training} During training the latent representation and encoding scheme of VASA, computation takes around two minutes using a pre-trained generative model on a reasonably high-end desktop PC. 
The end-to-end training VASA for the high-fidelity dataset at 1024$\times$1024 resolution is used 8 GPUs (V100). The training time for data with a resolution of 1024$\times$1024 is approximately 6 days, 4 days for data with a resolution of 512$\times$512 and 3 days for data with a resolution of 256$\times$256. Training is once-off, and could be reused for similar datasets, e.g., the model pre-trained on the FFHQ face dataset could be used for other human face datasets.
More evaluations on the complexity of the detection performance for our VASA-Defense are given in Section \ref{sec:performanceeva}.
\subsection{Defense evaluation against adversarial attacks}
\subsubsection{Evaluation of detection with Clean-Only consistency}
\begin{figure}[!htb]
	\setlength{\abovecaptionskip}{-0.05cm}
	\setlength{\belowcaptionskip}{-0.05cm}
	\includegraphics[width=2.0in]{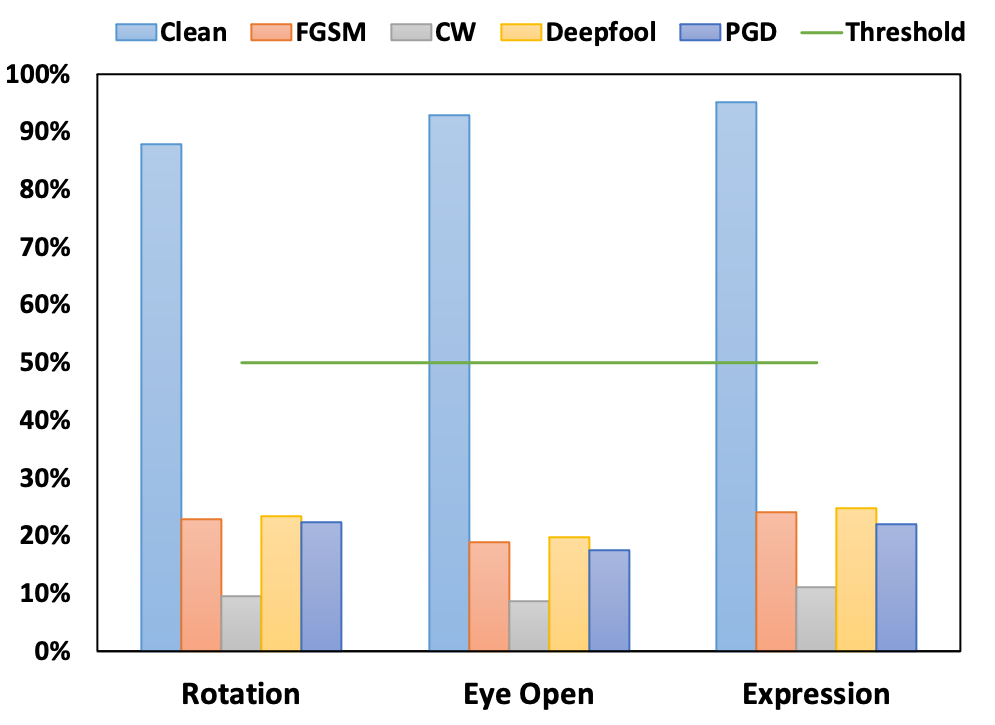}
	\centering
	\caption{Feasibility of threshold based detector on FFHQ face images, y is the averaged consistency value of edited copies under a single semantic feature axis.}
	\label{fig:threshold}
\end{figure}

This section evaluates the detection performance of the detection strategy proposed in \textit{Section 4.4.1 Detection with Clean-Only Consistency} against various unseen adversarial attacks. The default attacks, used in the classifier targeted attack setting in \textit{Section 5.1.3 Attack Settings}, include FGSM ($L_{\infty}$ distance, and $\epsilon = 0.005,0025,0.05$ for MNIST, LSUN cat and Face, respectively), Iterative attack ($L_{\infty}, \epsilon=0.01$ and $L_2, \epsilon=0.5$), Deepfool attack ($L_{\infty}$)  and C\&W attack ($L_{2}$). Other default parameters are provided in \textit{Appendix A. Architectures and Hyperparameters}. 
Consistency evaluation can be performed on the edited copies using latent shifting along with a variety of combinations of style axes in the latent space. We only consider one style axis for simplicity, and we observe that the performance is good even with only one style axis. 
First, we demonstrate the feasibility of using a simple threshold to distinguish clean and adversarial instances based on consistency values under a single-axis setting, as shown in Fig. \ref{fig:threshold}. 
\begin{figure}[!htb]
	\setlength{\abovecaptionskip}{-0.05cm}
	\setlength{\belowcaptionskip}{-0.05cm}
	\includegraphics[width=1.8in]{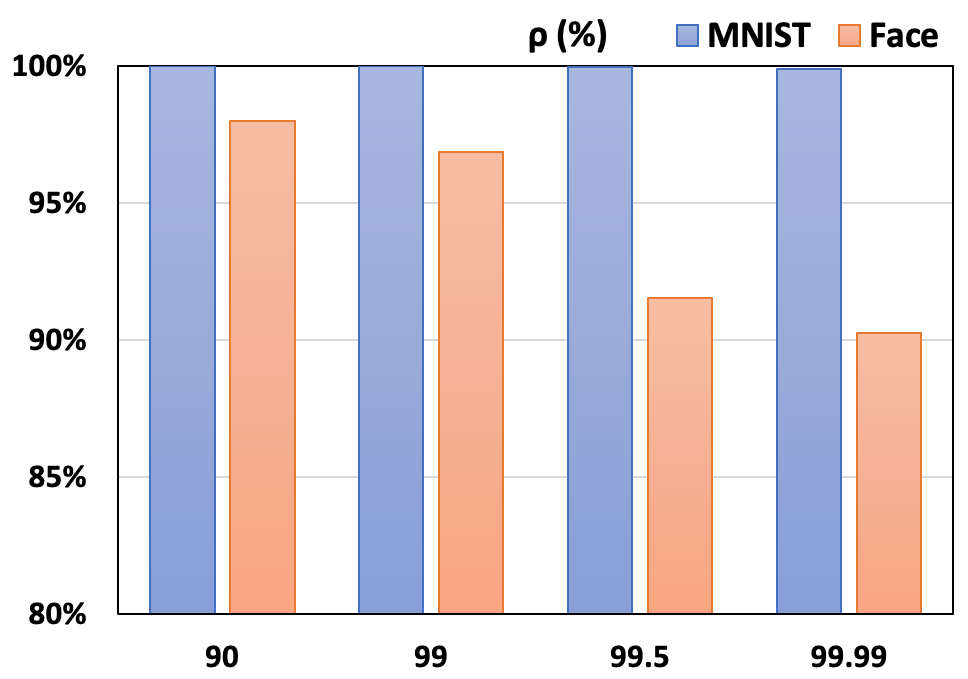}
	\centering
	\caption{\textcolor{black}{Detection accuracy when varying consistency threshold $\rho$ (\%) on MNIST and Face datasets under C\&W attack.}}
	\label{fig:rho}
\end{figure}
\begin{figure}[!htb]
	\setlength{\abovecaptionskip}{-0.05cm}
	\setlength{\belowcaptionskip}{-0.2cm}
	\includegraphics[width=3.0in]{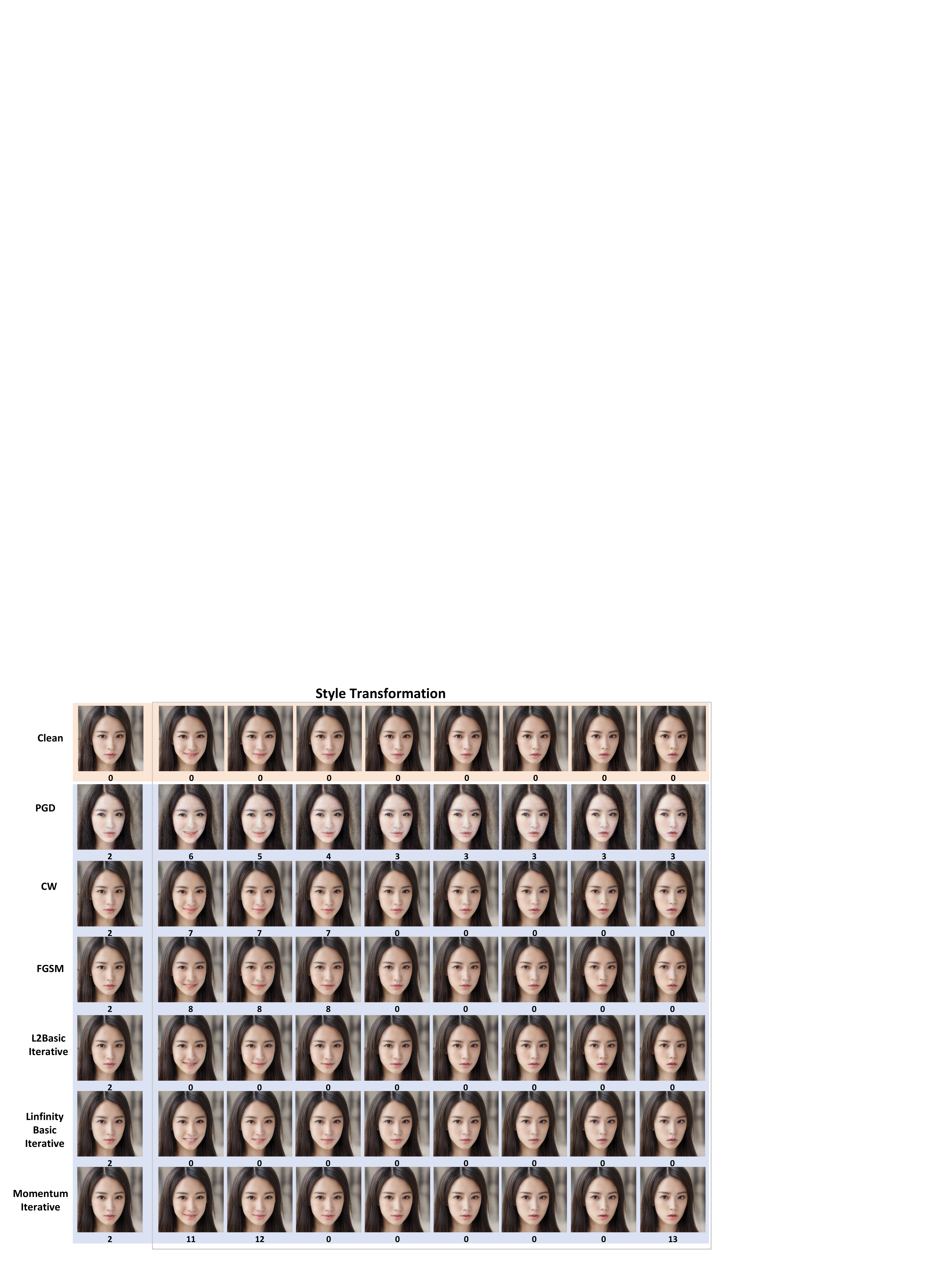}
	\centering
	\caption{Demonstration of VASA-defense on various adversarial attacks.}
	\label{fig:demoadv}
\end{figure}
\begin{figure*}[!htb]
	\includegraphics[width=5.0in]{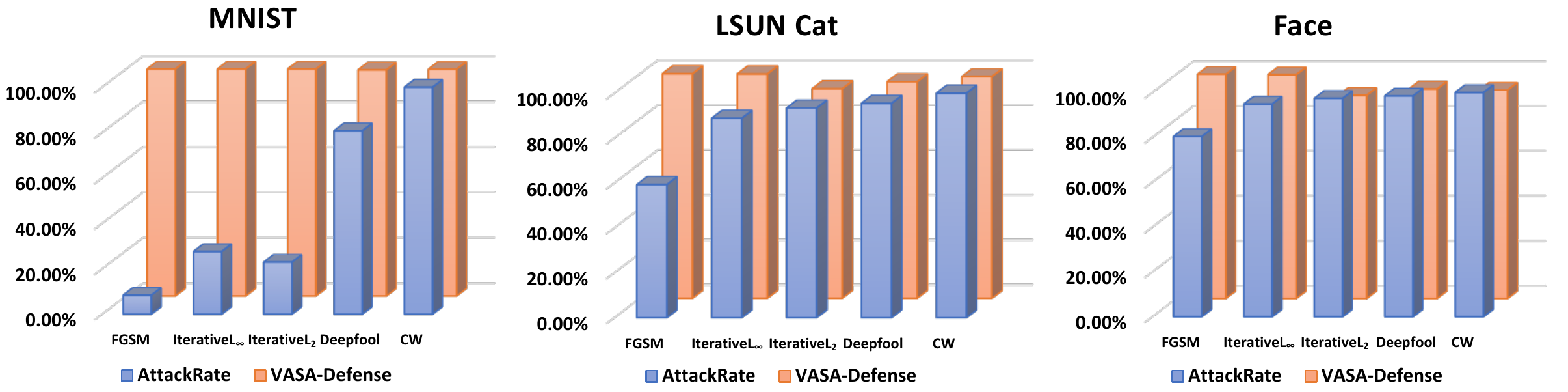}
	\centering
	\caption{Attacking rate and detecting accuracy on various adversarial attacks.}
	\label{fig:detectionacc}
\end{figure*}

We also demonstrate the detection accuracy when varying the consistency threshold $\rho$ to decide a specific consistency threshold using only a clean dataset. As shown in Fig. \ref{fig:rho}, detection accuracy increases as $\rho$ decreases (using a larger consistency value as threshold) for both MNIST and FFHQ Face datasets under C\&W attack. Even with the 99.99\% $\rho$ (almost the minimum consistency value for the Clean datasets), the detection accuracy for these two datasets could be achieved at around 99.9\% and 90.0\% detection, respectively, with a low false-positive rate (0.01\%) for distinguishing clean inputs. Therefore, it is possible to figure out a $\rho$ value to balance detection accuracy and false-positive rate (1-$\rho$). 

We further illustrate our intuition on a high-resolution face image (1024 $\times$ 1024) using the single style transformation and clean-only consistency threshold-based detection. As shown in Fig. \ref{fig:demoadv}, figures of the first column are original (underlying) images, where the first row is the clean image, and the rest of the rows are adversarial examples derived from various targeted attacks to misclassify the identity from class 0 to 2. The images in the second to the last column are edited copies with smiling style transformations applied to each underlying image in the first column. 
In the first column, we can see the classification consistency of the legitimate (clean) image. All edited copies have the same classification result as the underlying legitimate image.
The images from the second row to the last row demonstrate the classification consistency of different adversarial examples. In each row, the figures demonstrate that the classification results of the edited copies are almost all different from their underlying images. The perturbation of adversarial attacks drags the edited copies to unexpected predictions that differ from those of the underlying image.  
We also find that some adversarial examples can also be purified to the correct label (0) via the reconstruction of VASA, such as all the edited copies of basic iterative attacks in L2 and $L^{\infty}$ distances and parts of other attacks. The detection efficiency is also comparable to FFHQ using the LSUN cat (256*256) and car (512*512). We find that reconstructing the adversarial example to correct images for low-resolution images is more feasible. 

Next, we evaluate the performance of the \textit{Detection with Clean-Only Consistency} strategy on the CLE and ADV datasets, respectively. Adversarial examples are recognized as 1, while clean examples are recognized as 0. 
The adversarial detection accuracy (legend as ADV) is the proportion of adversarial instances in ADV to be recognized as adversarial, i.e., True-Positive ratio. The clean detection accuracy (legend as Clean) is the proportion of clean instances in CLE to be recognized as clean, i.e., True-Negative ratio. The overall detection accuracy (legend as Overall) is the proportion of all correctly detected instances in both ADV and CLE. 
The attacking rate of various attacks and the detection accuracy over ADV using MNIST, LSUN cat and Face are shown in Fig. \ref{fig:detectionacc}. 

\textcolor{black}{
The C\&W attack has been shown to be the most effective attack, with a success rate of $100\%$ for all types of images. 
The adversarial detection accuracy of VASA-Defense on ADV is near $100\%$ for MNIST on all the attacks, including the C \& W attack ($99.6\%$ adversarial detection accuracy with $99.9\%$ correctly recognized clean instances when $\rho=99.99\%$, i.e. rejection rate is 0.01\%). 
For high-resolution face images, the adversarial detection accuracy of VASA-Defense on ADV is above $90\%$ on all the attacks, including C\&W attack ($92\%$ adversarial detection accuracy with $99.3\%$ correctly recognized clean instances when $\rho=99.3\%$, i.e., the rejection rate is 0.07\%). 
}

\begin{figure}[!htb]
	\setlength{\abovecaptionskip}{-0.05cm}
	\setlength{\belowcaptionskip}{-0.05cm}
	\includegraphics[width=3.7in]{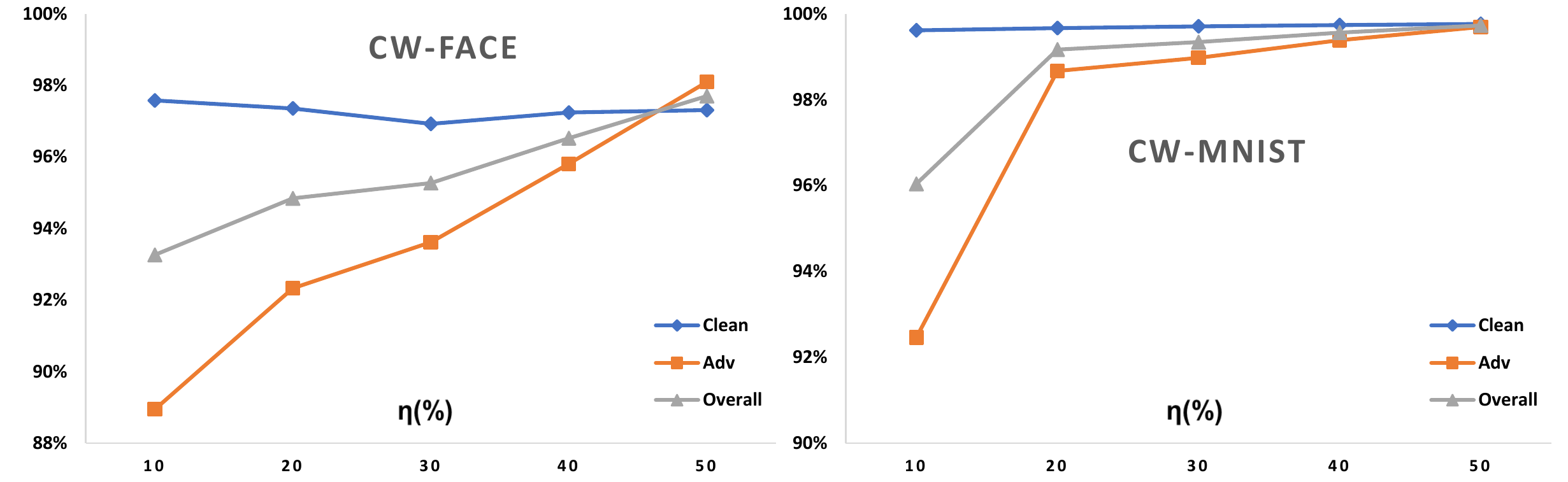}
	\centering
	\caption{Detecting performance for C\&W attack on MNIST and Face with varying shifting bounds.}
	\label{fig:eta}
\end{figure}

\textcolor{black}{
The bound of the latent shifting also has an impact on classification consistency as well as detection accuracy. 
Additionally, we examine how the bound of latent shifting affects consistency and detecting accuracy, as shown in Fig. \ref{fig:eta}. 
The bound is set as the $\eta$ (default 25\%) of the variance $\sigma_{style}$ on labelled data in our experiments, i.e., $\gamma \sim [-\eta \times \sigma_{style}, \eta \times \sigma_{style}] $ for Eq. \ref{eq:md}. 
We observe that even for a small shifting bound, e.g., $\eta=10\%$, it can detect more than $85\%$ adversarial examples, and retain more than $99\%$ clean instances correctly labeled for all types of images and evaluated attacks. 
As the bound increases (larger $\eta$), it will cause the classification consistency to drop for all the image sets (clean and adversarial), resulting in lower detection accuracy. However, the adversarial examples are significantly more sensitive to the style transformations compared with the clean image samples. 
}
\textcolor{black}{
Although the accuracy of adversarial detection increases as the consistency threshold is increased, the False Positive (FP) rate for clean instances also increases. It is necessary to investigate how to balance the adversarial detection accuracy and the FP rate. 
By leveraging the power of VASA's representation learning capabilities, we have observed that approximately $100\%$ clean images that are incorrectly identified as adversarial can be restored to the correct state via reconstruction. The reconstructed images retain some stochastic level features as a result of the learned noise via encoded latent code fine-tuning and noise embedding. 
}

\begin{figure}[!htb]
	\centering
	\setlength{\abovecaptionskip}{-0.05cm}
	\setlength{\belowcaptionskip}{-0.2cm}
	\includegraphics[width=2.0in]{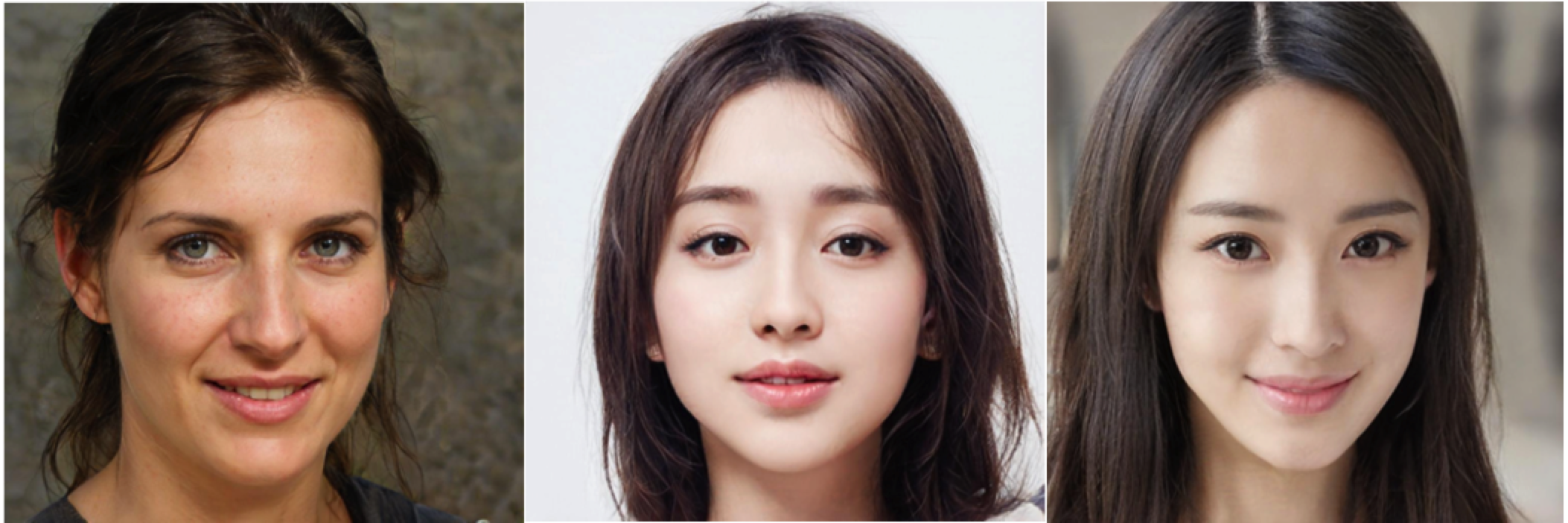}
	\centering
	\caption{Reconstruction quality illustration.}
	\label{fig:recquality}
\end{figure}

We illustrate the quality of the purified images on the high-resolution face image from FFHQ datasets through reconstruction by VASA in Fig. \ref{fig:recquality}. 
Samples reconstructed using VASA retain many stochastic features. 
Our noise embedding allows us to capture additional details during reconstruction, such as freckles, facial muscle lines, and hairlines, as shown in Fig. \ref{fig:recquality}. Therefore, our defense has a better generalization capability and can purify misjudged clean instances and some adversarial examples with small distortions. 
Our results confirm that VASA-defense is capable of efficiently recognizing adversarial examples. Additionally, VASA can effectively move the misjudged clean instances towards the normal manifold, further reducing the false-positive rate. The VASA-defense method can efficiently thwart adversarial examples while achieving a low FP rate.
\vspace{-4mm}
\begin{figure}[!htb]
	\includegraphics[width=3.5in]{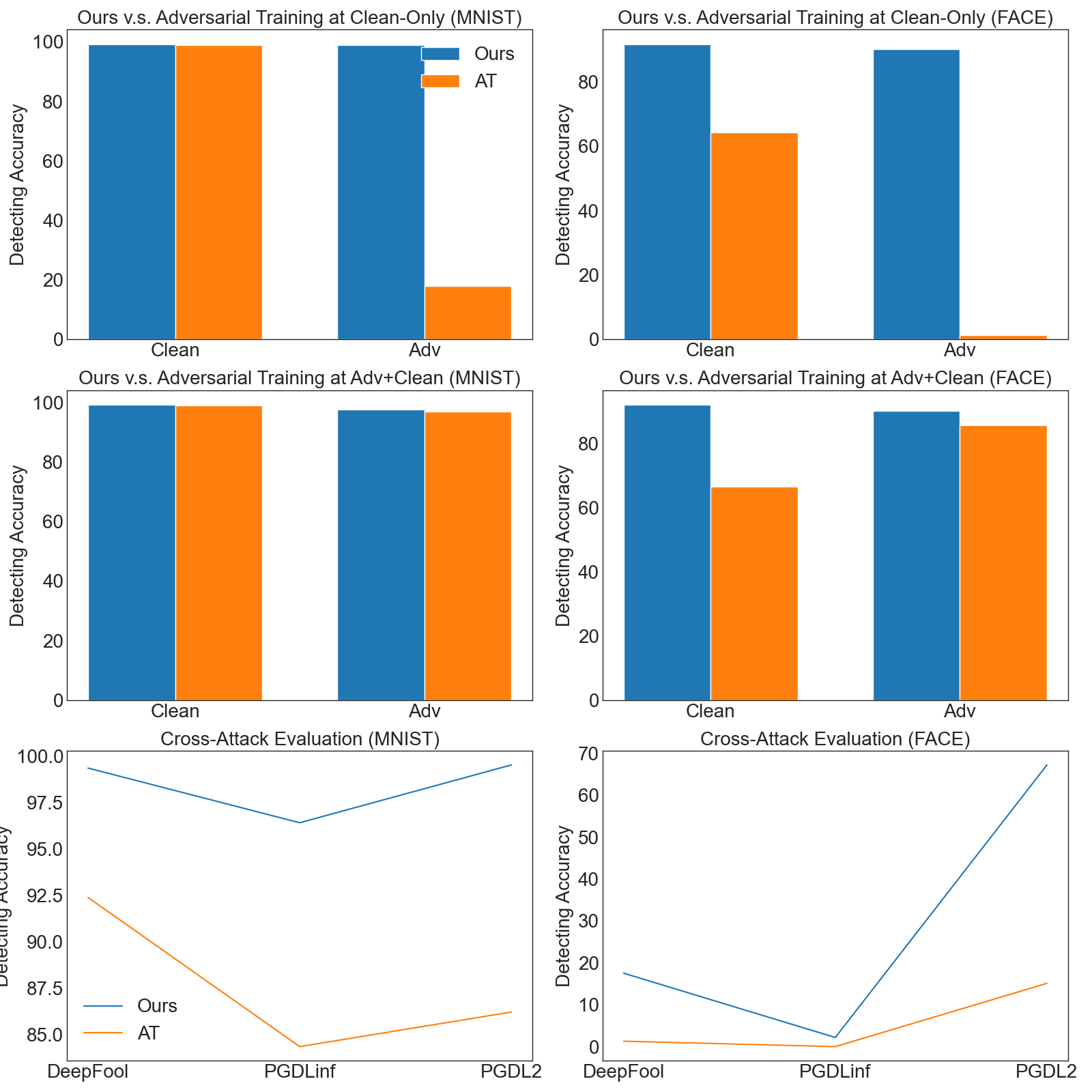}
	\centering
	\caption{\textcolor{black}{Detecting performance of comparison with ordinary adversarial training (AT) and cross-attack settings on MNIST and Face. Detection accuracy (\%) is on correctly recognized unseen adversarial samples.}}
	\label{fig:cross}
\end{figure}
\subsubsection{Comparison of detecting evaluations under Clean-Only and Adversarial Supervision scenarios}
\textcolor{black}{
Here, we compare the performance of detection strategies proposed in \textit{Section 4.4.1 Detection with Clean-Only Consistency} with the ordinary adversarial training \cite{goodfellow6572explaining}. The default setting is that our detector is only trained on clean instances. In contrast, ordinary adversarial training is trained using both clean instances and adversarial examples derived from FGSM with $\epsilon=0.3$ for MNIST and Face, under the classifier targeted attack setting in \textit{Section 5.1.3 Attack Settings}. It is assumed that the attacker is unaware of the detection mechanism. Other default parameters are provided in \textit{Appendix A. Architectures and Hyperparameters}. 
The detection performance is evaluated on both clean examples and adversarial examples. 
As demonstrated in Fig. \ref{fig:cross}, compared to ordinary adversarial learning methods, our defense achieves much better detection accuracy on adversarial examples. As shown in the first sub-figure of Fig. \ref{fig:cross}, our method achieves around 99\% detecting accuracy for MNIST data, as compared to 18\% for ordinary adversarial training approaches, and 90.16\% v.s. 11.87\% for Face data. 
}

\textcolor{black}{
Based on the assumption that computing resources are sufficient to construct a large number of adversarial examples, the evaluation of the defense is based on both clean and adversarial examples. 
According to Fig. \ref{fig:cross}, both the clean and adversarial testing datasets show that our defense outperforms conventional adversarial training.  
On clean examples, our defense demonstrates a higher degree of accuracy than the conventional adversarial training strategies (99.96\% vs. 99.09\% for MNIST, and 90.92\% vs. 66.34\% for Face). Also, our defense performs better on adversarial examples as well as traditional adversarial training (99.43\% vs. 97.01\% for MNIST and 92.16\% vs. 86.21\% for Face). 
This better performance of adversarial training is at the expense of requiring a large amount of memory and computing resources to generate a large number of adversarial examples. Furthermore, it is not possible to include all unknown attack samples in the adversarial training. 
On the other hand, our defense for the Clean-Only scenario also achieves fair defense performance without consuming any additional computing resources during adversarial example generation.
}
\subsubsection{Detecting evaluations with cross-attacks}
\label{sec:crossevaluation}
\textcolor{black}{
Another interesting aspect is the evaluation of the transferability of our defense to the scenario of cross-attacks. 
According to existing research \cite{goodfellow2014explaining,wen2020beneficial}, for ordinary adversarial training, a model can only defend against adversarial examples for which it has been trained. 
We compare the detection strategy proposed in \textit{Section 4.4.1 Detection with Clean-Only Consistency} with the ordinary adversarial training \cite{goodfellow6572explaining} under the classifier targeted attack setting and cross attack settings in Section 5.1.2. 
The training data is a combination of clean instances and their adversarial example counterparts derived from FGSM attack ($\epsilon=0.3$), and then detectors are evaluated against unseen attacks such as PGD ($L_2$ and $L_{\infty}$) and DeepFool. Note that the FGSM adversary examples in the training data are used to determine the consistency thresholds for our Clean-Only detector without additional training. }
The detailed results are given in the last row of Fig. \ref{fig:cross}.
We found that our defense derived from a Clean-Only scenario with thresholds determined through FGSM can not only defend well against FGSM attacks, but it can also be generalized to protect against even more complex attacks that have not been trained on. For instance, our detector has a transferable defending accuracy of 95\%
against both PGD Linf and L2 attacks from 99\% at FGSM for MNIST data, and almost 85\% from 90\% for Face data. In contrast, ordinary adversarial training would almost fail to transfer the defensive performance from FGSM to PGD attacks. 
Having such a transferable advantage allows our defense method to detect unknown attacks. 

\subsubsection{Overall evaluation with comparisons}
\textcolor{black}{
This section presents a comparison of our detection strategy proposed in \textit{Section 4.4.1 Detection with Clean-Only Consistency} against different adversarial attacks on the MNIST and FFHQ with state-of-the-art detection defenses, including MagNet, Defense-GAN, and FBGAN. We apply the classifier targeted attack setting in \textit{Section 5.1.3 Attack Settings}. 
}
As illustrated in Table \ref{tb:result}, the defense performance of VASA-Defense has stronger defense abilities than MagNet and Defense-GAN. 
The accuracy was obtained experimentally using the same test data. 
As shown, the performance of VASA-Defense exceeds that of MagNet on all adversarial attacks (DeepFool and C\&W). VASA-Defense also outperforms FB-GAN and Defense-GAN against FGSM attack. 

\textcolor{black}{
Furthermore, we compare the performance of our detection strategy proposed in \textit{Section 4.4.1 Detection with Clean-Only Consistency} with other image transformation based multi-point defense techniques, such as spatial rotation image transformation (RIT) \cite{tian2018detecting} and neural fingerprinting (NFP) \cite{dathathri2018detecting}. We consider the classifier a targeted attack setting. 
We compare the performances of our approach and RIT and NFP under the cross-adversary benchmark in Fig. \ref{fig:imagetrans}. Specifically, we allow supervision for the training of RIT and NFP, in which a set of clean instances and associated adversarial examples from C\&W attack are used as the training data for the detector, while the DeepFool attack is used to test the detection accuracy. However, we only use the clean data set for VASA-Defense to determine a threshold for the detector. 
}
Fig. \ref{fig:imagetrans} demonstrates that our VASA-Defense outperforms other multi-point image transformation-based detection methods for both datasets. 
Note that our approach achieves such good performance without additional information about the adversary and further supervised training.  
Therefore, VASA-Defense is capable of detecting new attacks without prior knowledge of those attacks.  
\begin{table}[!htb]
	\centering
	\setlength{\abovecaptionskip}{-0.03cm}
	\setlength{\belowcaptionskip}{-0.1cm}
	\scriptsize
	\caption{Classifier accuracy ($\%$, MNIST/FFHQ) under various attack and defense methods.}
	\label{tab:my-table}
\begin{tabular}{l|l|l|l|l|l}
\hline
         & No-defense & VASA      & Def-GAN & MagNet    & FBGAN     \\ \hline
FGSM     & 91.2/25.6  & 100/99.9  & 83.2/32.1   & 74.6/59.2 & 80.4/33.7 \\
C\&W       & 0/0        & 99.9/92.9 & 80.1/28.2   & 19.6/40.5 & 90.8/35.5 \\
DeepFool & 19.1/2.1   & 99.6/93.8 & 81.1/30.1   & 49.4/53.4 & -         \\ \hline
\end{tabular}
\label{tb:result}
\end{table}
\begin{figure}[!htb]
	\setlength{\abovecaptionskip}{-0.05cm}
	\setlength{\belowcaptionskip}{-0.05cm}
	\includegraphics[width=2.4in]{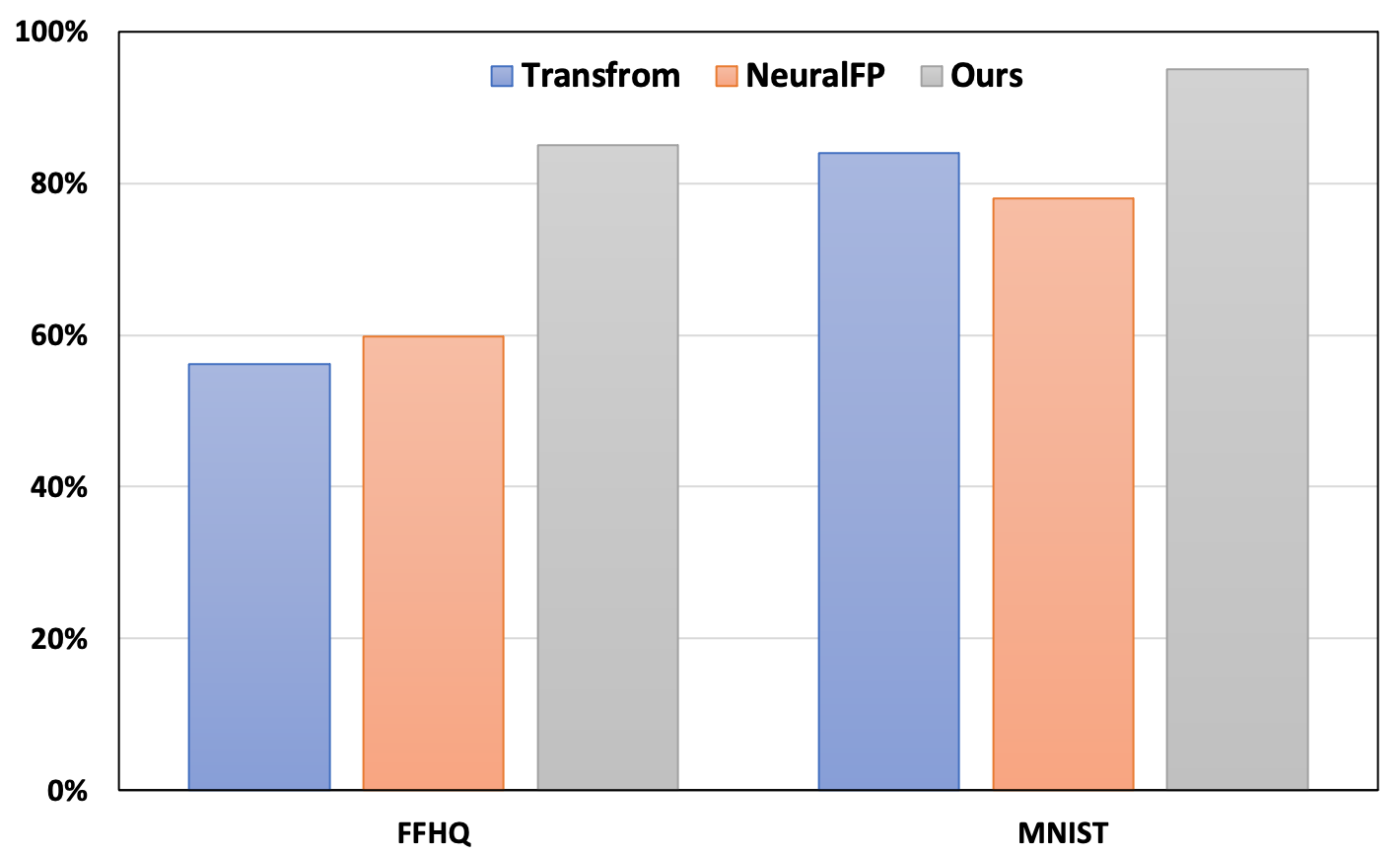}
	\centering
	\caption{\textcolor{black}{Detection accuracy ($\%$, MNIST/FFHQ) under cross-adversary benchmark for NFP \cite{dathathri2018detecting}, RIT \cite{tian2018detecting}  and our VASA-Defense (Clean-Only).}}
	\label{fig:imagetrans}
\end{figure}
\subsubsection{Dependence between the number of copies and detection}
\label{sec:performanceeva}
\begin{figure}[!htb]
	\setlength{\abovecaptionskip}{-0.05cm}
	\setlength{\belowcaptionskip}{-0.05cm}
	\includegraphics[width=3.6in]{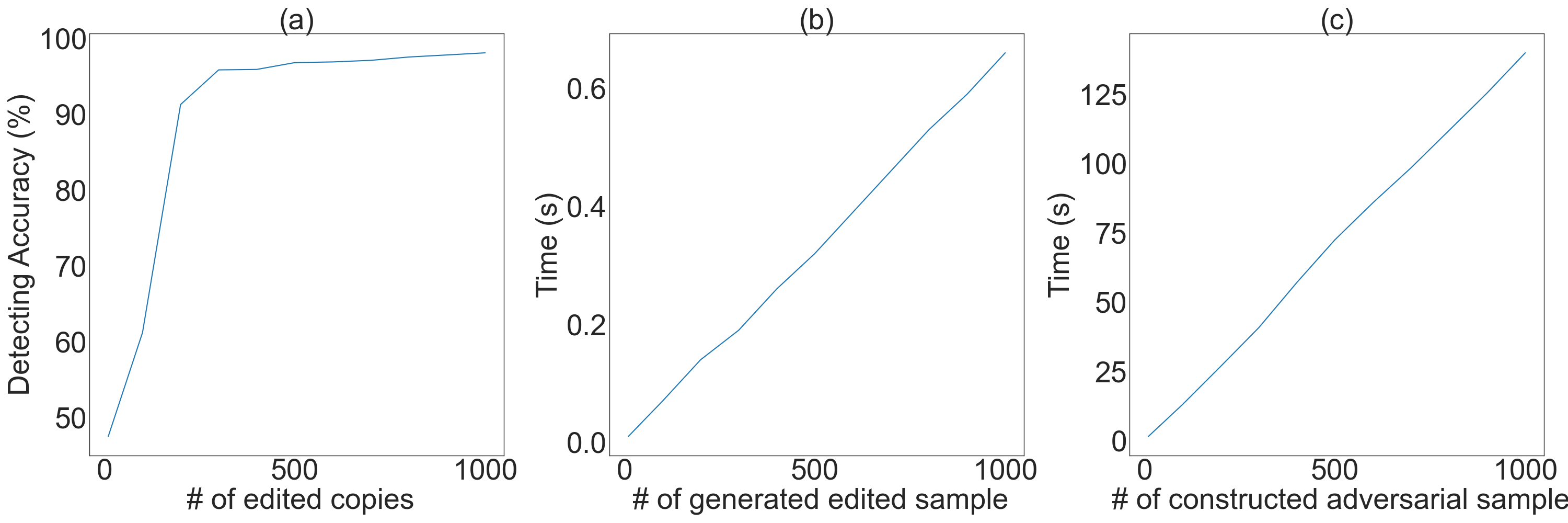}
	\centering
	\caption{Detecting performance and time cost for PGD Linf ($\epsilon=0.3$) attack on Face (224*224 resolution). (a) Accuracy v.s. number of edited copies for each sample. (b) Time v.s. number of generated edited sample. (c) Time v.s. number of constructed adversarial sample.}
	\label{res:number}
\end{figure}

\textcolor{black}{
Here, we evaluate the correlation between number of copies and detecting performance as well as the time complexity involved. 
We apply the detection strategy proposed in \textit{Section 4.4.1 Detection with Clean-Only Consistency} on Face dataset. The default attacks are under the classifier targeted attack setting in \textit{Section 5.1.3 Attack Settings}, including FGSM ($L_{\infty}$ distance, and $\epsilon =0.05$ for Face, respectively), Iterative attack ($L_{\infty}, \epsilon=0.01$ and $L_2, \epsilon=0.5$), Deepfool attack ($L_{\infty}$)  and C\&W attack ($L_{2}$). 
Other default parameters are provided in \textit{Appendix A. Architectures and Hyperparameters}.} 
Regarding the number of edited copies and the detection performance, Fig. \ref{res:number} illustrates the relationship between the number of edited copies generated by VASE and the detection performance. 
The first subfigure of Fig. \ref{res:number} illustrates how our method allows users to achieve a balance between accuracy and the number of edited copies. The reported detection accuracy is averaged among various attacks. In general, 200 copies are sufficient to achieve moderate robustness by detecting $\geq 90\%$ adversarial examples.

Furthermore, we illustrate the time cost of our method in comparison to conventional adversarial training on an end-user laptop with limited computing power (16 G i7 CPU and 8 GB GPU), in terms of training data preparation, detector training, and detector inference. 
\textit{(1) Training data preparation.} In general, the preparation of samples for detector training is the most time-consuming step. 
We first investigate the time to prepare the training samples for the detector, i.e., adversarial samples for adversarial training, and edited copies for our defense. 
A greater number of adversarial examples is required for adversarial training, which is a resource-intensive process, particularly when using high-fidelity images. 
The last two subfigures of Fig. \ref{res:number} demonstrate the times involved in generating 1000 samples of our edited copies and adversarial copies (PGD attacks) on Face images. As shown, our method only requires 0.5\% of the time cost for preparing training instances compared to adversarial training. Generally, the time to produce one adversarial example for one clean 224*224 resolution instance could generate more than 200 edited copies for one instance. 
Our method differs from conventional adversarial training in that the manipulation takes place in a low-dimensional latent space instead of a high-dimensional pixel space. The generation time increases with the number of generations in both scenarios, but the time required to achieve moderate defensive performance is much less with our method. 
\textit{(2) Detector Training. }During the training of the detector, adversarial training requires the use of a neural network as a classifier, using both clean instances and adversarial instances in the input pixel space, which is time-consuming, especially for high-dimensional inputs. In contrast, the detector for our defense is based on the consistency threshold, without introducing any extra time cost. 
Consequently, our defense yields significant computational savings during the training process, especially when dealing with large datasets with high-resolution images and new attacks. 
In the case of end-users with modest computational resources, our approach may be able to allow the system to attain good robustness against adversarial examples at a negligible additional cost. 
\textit{(3) Detector Inference. } In the inference stage for detection, for adversarial training, inference time is used for the feedforward procedure of the trained classifier, which is often very small, such as 0.026 seconds for one Face image. For our defense, a set of edited copies are required to be generated to produce the consistency vector via the feedforward procedure of the trained generator. As demonstrated in the first subfigure of Fig. \ref{res:number}, 200 edited copies could achieve more than 90\% detection accuracy, which takes around 0.12s for one Face image. Though the inference time for our defense is slightly longer than the inference time for the classifier from adversarial training, it is acceptable for practice. In order to enhance practicality, the edited copy generation could be conducted in parallel. 

\subsubsection{Detecting evaluations under White-box attacks}
\begin{figure}[!htb]
	\setlength{\abovecaptionskip}{-0.05cm}
	\setlength{\belowcaptionskip}{-0.05cm}
	\includegraphics[width=3.6in]{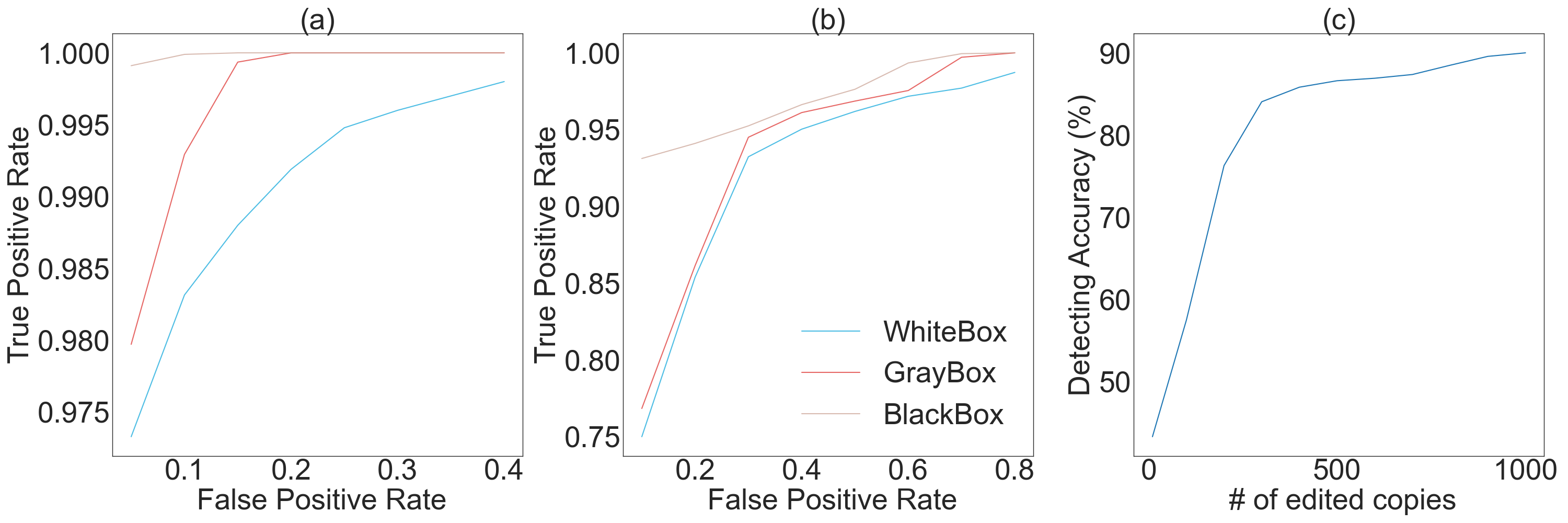}
	\centering
	\caption{Detecting performance for under various attacking settings. (a) FP v.s. TP for MNIST. (b) FP v.s. TP for FACE. (c) Accuracy v.s. number of edited copies for adaptive attack.}
	\label{res:box}
\end{figure}

\textcolor{black}{
In previous experiments, we have demonstrated the efficiency of our detection strategy proposed in Section 4.4.1 Detection with Clean-Only Consistency under the classifier targeted attack setting. In this section, we further evaluate the performance of our detection approach, which combines strategies proposed in Section 4.4.2 Detection with adversarial supervision and Section 4.4.2 Detection against adaptive attacks with White-Box knowledge about both the classifier and detector, under the detector targeted and classifier+detector targeted attack settings in Section 5.1.3 Attack Settings, respectively. Default parameters are provided in \textit{Appendix A. Architectures and Hyperparameters}.} 
Fig. \ref{res:box} illustrates that the attack with white-box knowledge about classifier+detector is, as expected, the most effective attack among the black box and gray box attacks. 
Our detection against classifier+detector white-box attacks can detect adversarial examples that have been created by attacking both the classifier and detector on the MNIST dataset. This results in more than 98\% detecting accuracy with less than 10\% False Positive rates. For the Face results, the classifier+detector white-box attack is also the most efficient approach in the second subfigure. The detection against classifier+detector white-box attacks can also recognize adversarial examples derived from targeting both the classifier and detector on the Face dataset, resulting in an overall detection accuracy of over 80 percent with a false positive rate of less than 20 percent. 

\textcolor{black}{
We also evaluate the attack success rate of the Expectation over Transformation (EoT) \cite{athalye2018synthesizing} based adaptive attack in \cite{tramer2020adaptive,athalye2018obfuscated} against the aforementioned our defense on Face dataset. 
With our defense, the advanced adaptive attacker is only able to achieve under 20\% attack success rate. The attacker can achieve about $99\%$ attack success rate against unguarded models under the same perturbation budget. Besides, the memory cost also increases 4 times to bypass our detection.
Additionally, more perturbations are required to create adversarial inputs that bypass our defenses. It is demonstrated that even when adaptive attackers are aware of our defense, our defense can still increase the attacker's effort towards invulnerability, e.g., by requiring greater perturbation budgets and computing resources to cause the misclassification. 
Besides, the detection accuracy of adaptive attacks increases when the number of edited copies is increased. It is because our defense introduces more uncertainty when the sampling points on latent codes become dense for the style transformation. 
As a result, they are unable to fully capture the non-deterministic transformation procedures with the gradient descent method. 
}
\section{Discussion}
\textcolor{black}{
In this section, we discuss how our defense could address the bottleneck issues faced by existing detection systems and handle strong adversaries. 
} 
\textcolor{black}{
A key reason for bypassing current generative model-based detection methods is that detection is carried out by evaluating pixel-wise differences between each original input and its single reconstruction, i.e., reconstruction error. Considering the deterministic generative model for reconstruction, the reconstruction error could also be deterministic. As both the adversary and detector operate in pixel space from the perspective of one instance based on the same deterministic scheme, one side is always able to defeat the other, as a Min-Max zero sum game \cite{yuan2019adversarial,bose2020adversarial}. Therefore, there will always be strong attacks to undermine the current defense strategy, for example by introducing minimal pixel perturbations \cite{carlini2017towards}. 
Alternatively, consistency-based detection approaches recognize the differences in patterns between benign and adversarial instances by performing certain spatial transformations on the input images. Even so, spatial transformations such as rotation and scaling are also deterministic and cannot be applied automatically. Additionally, these approaches tend to be attack-specific and are only valid for a specific dataset. 
}

\textcolor{black}{
In our VASA-Defense system, we address these bottleneck issues from three different perspectives in order to achieve a more competitive detection, as demonstrated in the experiments. 
(1) Our defense goes beyond an examination of a single instance to a consideration of the entire process. Information sources are expanded from a single point source to multi-point sources, resulting in robust detection against a variety of attacks (even unknown ones). It is demonstrated that our Clean-Only defense could achieve more than 90\% detection success on all evaluated datasets against various attacks with a tiny false positive rate (around 0.01), despite the fact that the evaluated attacks are unknown to the defense. 
(2) Our Clean-Only defense could be free of cost-intensive training procedures without incorporating adversarial counterparts. Our experiments demonstrate that the simple consistency threshold-based detector could achieve efficient detection accuracy against unseen attacks under the classifier targeted attack settings, using only clean instances. 
(3) We incorporate latent randomness into the stochastic non-linear transformation.  
The reason for the failure of existing defenses with randomization is that randomness is added in pixel space, while there is a deterministic connection between the gradient information and pixel. Strong attacks, such as EoT-based attacks, can approximate the gradient through the transformation (even with randomization) in the pixel space in order to bypass these defenses. 
However, our defense breaks the connection between transformation and gradient to defeat such strong attacks. 
First, stochastic variation is introduced through the addition of noise to the generator model for style transformation, introducing randomness and uncertainty into the image transformation. 
Second, we introduce further randomness into the latent space to manipulate transformation via non-linear mapping between latent and pixel space in a black-box manner. 
Therefore, such uncertainty and randomness could reduce the feasibility of a strong adversary to fully capture the non-deterministic transformation procedures from the pixel space for gradient-based optimization.
}

\textcolor{black}{
It is demonstrated that our combined defense could achieve an overall detection accuracy of over 80 percent on all test datasets against classifier+detector white-box attacks (adaptive attacks). We also demonstrate that the advanced EoT-based adaptive attacker can only achieve under $20\%$ attack success rates against the models with our defense. 
To further improve the robustness against such classifier+detector white-box attacks or adaptive attacks, randomness could be applied to generate a great number and large diversity of VASA candidates and randomly select one of these VASAs for each defensive device for every session, every test set, or even every test example. 
}
\section{Conclusion}
We propose VASA-Defense, an effective defense capable of detecting a wide range of state-of-the-art adversarial attacks. 
To conduct a style transformation on high-resolution images, we propose the VASA to encode images into disentangled latent codes which reveal hierarchical styles. A style editing axis is then identified in the latent space to reveal correspondences between the styles and latent codes. 
The latent codes are shifted along the latent style axis to generate a set of edited copies as a result of style transformations. 
It is possible to detect suspicious input based on the consistency of classification results for an image and its edited copies with style transformations. 

\textcolor{black}{
The VASA-Defense has demonstrated high detection accuracy against state-of-the-art attacks, including high fidelity images, providing empirical evidence that our assumptions are likely to be accurate. 
Additionally, VASA's strong reconstruction ability allowed it to purify misjudged clean instances with good quality and further reduce false-positive errors. 
Our defense only uses consistency evaluation on edited copies, however, the reconstruction error can also be computed at the same time. By having such pixel-level reconstruction errors, one would be able to deal with adversarial examples with large distortions. In the future, we will combine consistency-based detection and reconstruction error-based evaluation.
The experiments demonstrate that VASA can detect a wide range of adversarial examples with a high degree of accuracy. However, without stronger justification or proof, we cannot dismiss the possibility that there are future, currently unknown attacks that may thwart VASA's detection. 
}
\bibliographystyle{IEEEtran}
\bibliography{sample-sigconf}
\renewcommand{\baselinestretch}{0.8}
\appendix
\section*{A. Architectures and Hyperparameters}
\label{sec:hyper}
\textcolor{black}{
\textit{\textbf{A.1 Victim Models and Attacks. }}
Victim models for MNIST use the setting in \cite{carlini2017towards}.
The classifier is a neural network that consists of 9 layers, two 2*[Convolution+ReLU]+MaxPooling, and two FullyConnected+ReLu before SoftMax, with an accuracy of  $99.4\%$. The optimizer is SGD with momentum, and the mini-batch size is 128. Weight decay is 0.5, the epoch is 50, and the learning rate is 0.1. 
For the LSUN and FFHQ, victim classifier is based on the setting in \cite{simonyan2014very} with an accuracy of $98.2\%$ and $ 94.7 \%$ for identification recognition. 
The classifier is fine-tuned on the pre-trained VGG-16 \cite{simonyan2014very}, consisting of 16 convolutional and fully connected layers that mostly have 3×3 filters.
}

\textcolor{black}{
The default hyperparameters for the attacks are given as follows. 
For the fast gradient sign method (e.g., FGSM \cite{kurakin2016adversarial}), we use $L^{\infty}$ distance ($\epsilon=0.005,0025,0.05$ for MNIST, LSUN and FFHQ). 
For Projected Gradient Descent Attack \cite{madry2017towards} and  Iterative attacks \cite{kurakin2016adversarial,dong2018boosting}, we apply the L2 ($\epsilon=0.5$) and  $L^{\infty}$ ($\epsilon=0.01$) distances. 
For DeepFool \cite{moosavi2016deepfool}, we apply the $L^{\infty}$ distance. 
For C\&W attack \cite{carlini2017towards}, we apply the $L^{2}$ distance. 
The default steps and step size settings for MNIST are (200, 0.1) for $L^2$ distance and (200, 0.01) for $L^{\infty}$. 
The default steps and step size settings for LSUN and FFHQ are (2000, 0.05) for $L^2$ distance and (2000, 0.005) for $L^{\infty}$. 
}

\textcolor{black}{
As the target is the victim classifier C for the classifier targeted attack, adversarial examples x' for training $L^{2}$ and $L^{\infty}$ models are both optimized using the method of untargeted attack in C\&W \cite{carlini2017towards} attack by linking the x' to the logit outputs of C via Eq. \ref{eq:black}.
In detector White-box and classifier+detector white-box attacks, the detectors are also the targets. Given an input x, we assume the adversary could obtain its consistency vector $textbf{v}_x$. The adversarial sample $x'$ is calculated by minimizing the loss for the logit of the detector in Eq. \ref{eq:gray} and \ref{eq:white} by the normalized steepest descent in Projected Gradient Descent Attack \cite{madry2017towards} attacks. 
}

\textcolor{black}{
\textit{\textbf{A.2 Detectors. }}
The input for our detector is an m-dimensional consistency vector, where m is the number of augmented edited copies via style transformation, and each element is 0 or 1, details in Section \ref{sec:supervise}. 
Therefore, all base detectors will be implemented with a simple neural network composed of two Convolutional+MaxPooling layers, each with 32 and 64 filters, followed by two fully connected layers of size 1024. The optimizer is SGD with momentum, and the mini-batch size is 400. The weight decay is 0.5, the epoch is 100, and the learning rate is 0.1. 
}

\begin{IEEEbiographynophoto}{Shuo Wang}
\scriptsize
Dr. Shuo Wang is a Research Scientist in the CSIRO's Data61 and Cybersecurity CRC. His Ph.D. was in the University of Melbourne. His main research interests include the areas of: 
Trustworthy AI, security and privacy in systems, networking, and databases.
\end{IEEEbiographynophoto}
\begin{IEEEbiographynophoto}{Surya Nepal}
\scriptsize
Dr. Surya Nepal is a Senior Principal Research Scientist working on trust and security aspects of Web Services at CSIRO's Data61. 
\end{IEEEbiographynophoto}
\begin{IEEEbiographynophoto}{Alsharif Abuadbba}
\scriptsize
Dr. Alsharif Abuadbba is Senior Research Scientist at CSIRO's Data61, Australia. 
Alsharif has joined Data61 Distributed System Security group early 2019 as a Research Scientist and Cybersecurity CRC fellow.
\end{IEEEbiographynophoto}
\begin{IEEEbiographynophoto}{Carsten Rudolph}
\scriptsize
Carsten Rudolph is an Associate Professor of the Faculty of IT at Monash University, and Director of the Oceania Cyber Security Centre OCSC. 
\end{IEEEbiographynophoto}
\begin{IEEEbiographynophoto}{Marthie Grobler}
\scriptsize
Dr. Marthie Grobler is Principal Research Scientist at CSIRO's Data61 and leads the human-centric cybersecurity team.
\end{IEEEbiographynophoto}

\end{document}